\documentclass[10pt,twocolumn,letterpaper]{article}

\usepackage{cvpr}              

\usepackage{overpic}
\usepackage{enumitem} %
\usepackage{overpic} %
\usepackage{color}
\usepackage{xspace}
\usepackage{shortbold}
\usepackage{verbatim}
\usepackage{microtype}

\definecolor{turquoise}{cmyk}{0.65,0,0.1,0.3}
\definecolor{purple}{rgb}{0.65,0,0.65}
\definecolor{dark_green}{rgb}{0, 0.5, 0}
\definecolor{red}{rgb}{0.8, 0.2, 0.2}
\definecolor{darkred}{rgb}{0.6, 0.1, 0.05}
\definecolor{blueish}{rgb}{0.0, 0.3, .6}
\definecolor{light_gray}{rgb}{0.7, 0.7, .7}
\definecolor{pink}{rgb}{1, 0, 1}
\definecolor{greyblue}{rgb}{0.25, 0.25, 1}
\definecolor{msorange}{rgb}{0.93,0.49,0.19}
\definecolor{msblue}{rgb}{0.33,0.56,0.76}

\definecolor{darkyellow}{rgb}{0.796,0.901,0}
\newcommand{\duster}{DUSt3R\xspace}
\newcommand{\method}{Dyna\-DUSt3R\xspace}
\newcommand{\dataset}{Stereo4D\xspace}
\newcommand{\monster}{MonST3R\xspace}

\newcommand{\bfpar}[1]{{\vspace{1.2mm} \par \noindent \bf{{#1}}}}

\definecolor{BlueCam}{rgb}{0.10196, 0.52157, 1.0}
\definecolor{GreenCam}{rgb}{0, 0.5, 0}
\definecolor{PurpleCam}{rgb}{0.74,0.415,0.831}

\newcommand{\loss}[1]{\mathcal{L}_\mathsf{#1}}

\newcommand{\Fig}[1]{Fig.~\ref{fig:#1}}

\newcommand{\Tab}[1]{Tab.~\ref{tab:#1}}

\newcommand{\Sec}[1]{Sec.~\ref{sec:#1}}

\usepackage{blindtext}

\renewcommand{\paragraph}[1]{\vspace{1em}\noindent\textbf{#1}.}

\makeatletter
\DeclareRobustCommand\onedot{\futurelet\@let@token\@onedot}
\def\@onedot{\ifx\@let@token.\else.\null\fi\xspace}

\def\etal{\emph{et al}\onedot}
\makeatother

\definecolor{cvprblue}{rgb}{0.21,0.49,0.74}
\usepackage[pagebackref,breaklinks,colorlinks,allcolors=cvprblue]{hyperref}

\def\paperID{6709} 
\def\confName{CVPR}
\def\confYear{2025}

\title{\dataset: Learning How Things Move in 3D from Internet Stereo Videos}

\author{
Linyi Jin$^{1,2}$\qquad
Richard Tucker$^1$\qquad
Zhengqi Li$^1$\qquad
David Fouhey$^3$\\
Noah Snavely$^{1*}$\qquad
Aleksander Ho\l{}yński$^{1*}$
\\[0.5em]
$^1$Google DeepMind \ \ \
$^2$University of Michigan \ \ \ 
$^3$New York University \ \ \ 
$^*$equal contribution\\ \\
}

\begin{document}
\twocolumn[{%
\renewcommand\twocolumn[1][]{#1}%
\maketitle
\vspace{-2em}
 \includegraphics[width=\textwidth]{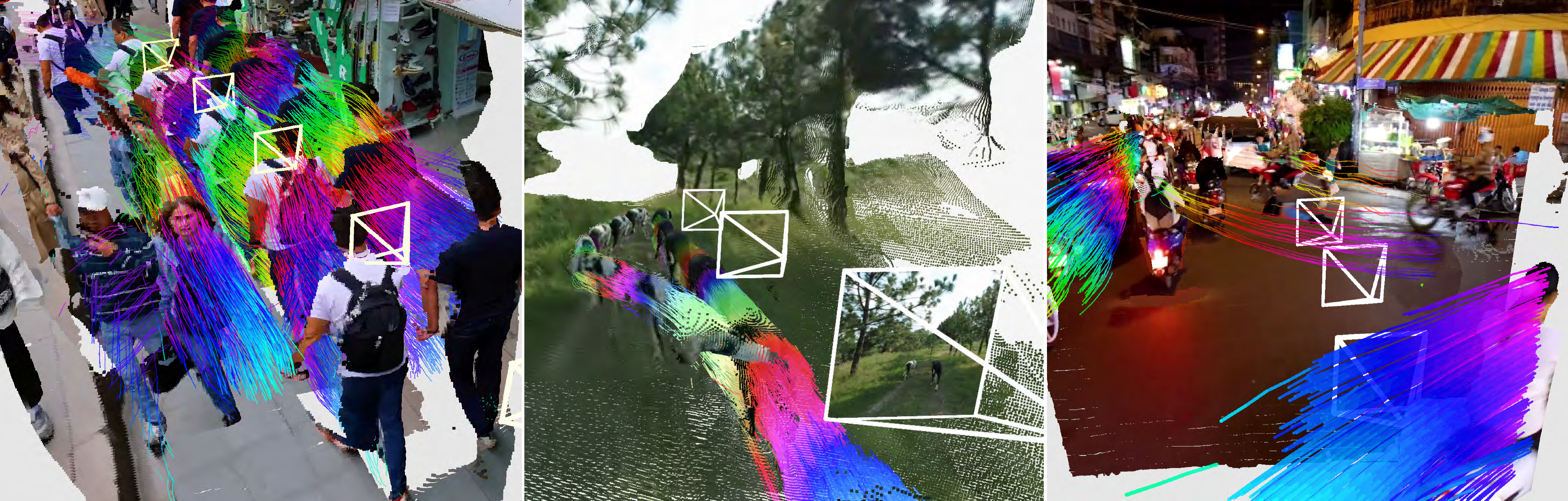}
\captionof{figure}{There is currently no scalable source of data for real-world, ground truth 3D motion paired with video. 
We present a framework for mining such data from existing stereoscopic videos on the Internet, in the form of 3D point clouds with long-range world-space trajectories. Our framework fuses and filters camera poses, dense depth maps, and 2D motion trajectories to produce high-quality, pseudo-metric point clouds with long-term 3D motion trajectories, pictured above, for hundreds of thousands of video clips. We show how this data is useful in learning a model that reasons about both 3D shape and motion in imagery.\\ }
\label{fig:teaser}
}]
\begin{abstract}
Learning to understand dynamic 3D scenes from imagery is crucial for applications ranging from robotics to scene reconstruction. Yet, unlike other problems where large-scale supervised training has enabled rapid progress, directly supervising methods for recovering 3D motion remains challenging due to the fundamental difficulty of obtaining ground truth annotations. 
We present a system for mining high-quality 4D reconstructions from internet stereoscopic, wide-angle videos. Our system fuses and filters the outputs of camera pose estimation, stereo depth estimation, and temporal tracking methods into high-quality dynamic 3D reconstructions. We use this method to generate large-scale data in the form of world-consistent, pseudo-metric 3D point clouds with long-term motion trajectories. We demonstrate the utility of this data by training a variant of \duster to predict structure and 3D motion from real-world image pairs, showing that training on our reconstructed data enables generalization to diverse real-world scenes. Project page and data at
\url{https://stereo4d.github.io}
\end{abstract}
    
\section{Introduction}
\label{sec:intro}

Simultaneously predicting and understanding geometry and motion---that is, dynamic 3D content---from images is a fundamental building block for computer vision, with applications ranging from robotic interaction and scene reconstruction to novel view synthesis of dynamic scenes. 
Recent work has made remarkable progress in predicting static 3D structure from images~\citep{wang2024dust3r,yang2024depth,bochkovskii2024depth}, but modeling real-world 3D motion---people gesturing, balls bouncing, leaves rustling in the wind---remains a critical unsolved challenge for building truly general models of the visual world.

Recent breakthroughs in AI, from 
large language models~\cite{achiam2023gpt,team2023gemini} to 
image generation~\cite{polyak2024movie} and 
static 3D reconstruction~\cite{wang2024dust3r,bochkovskii2024depth,yang2024depth}, 
demonstrate a consistent pattern: large amounts of high-quality, realistic training data and scalable architectures enable dramatic performance improvements. 
In the realm of 3D reasoning, prior works~\cite{li2018megadepth, depthanything, ranftl2020towards, ranftl2021vision, wang2024dust3r} have shown the value of large-scale training data for strong zero-shot generalization within single-view or two-view static scene settings.
But applying this same formula to \emph{dynamic} 3D scenes (i.e., moving 3D structure) requires a corresponding large-scale dataset consisting of diverse visual content paired with corresponding ground-truth 3D motion trajectories. 
Obtaining such data presents unique challenges. While there are synthetic datasets~\cite{zheng2023point,butler2012naturalistic,dosovitskiy2015flownet,greff2021kubric}, these often fail to capture the distribution of real-world content and the nuanced patterns of real-world motion. 
Traditional approaches to gathering real motion data, such as motion capture systems or multi-view camera arrays~\cite{Joo_2015_ICCV,Grauman_2024_CVPR,kirschstein2023nersemble,isik2023humanrf}
are accurate, but difficult to scale and limited in the diversity of scenes they can capture. 

We identify online stereoscopic fisheye videos (often referred to as VR180 videos) as an untapped source of such data. 
These videos, designed to capture immersive VR experiences, provide wide field-of-view stereo imagery with a standardized stereo baseline. We present a pipeline that carefully combines state-of-the-art methods for stereo depth estimation and video tracking along with
structure-from-motion methods optimized for dynamic scenes. By combining our system with careful filtering and quality control, we show that we can extract over 100K video sequences, each containing high-quality 3D point clouds with per-point long-term trajectories (see \Fig{teaser}), as well as all other intermediate quantities: depth maps, camera poses, images, and 2D correspondences. 
We additionally show the utility of the dataset by training \emph{\method}, an extension to \duster that can predict high-quality 3D structure {\it and} motion from challenging image pairs. 

Our contributions include: (1) a framework to obtain real-world, dynamic, and pseudo-metric 4D reconstructions and camera poses at scale from online videos; and (2) \method, a method that, given a pair of video frames, predicts a pair of 3D point clouds and the corresponding 3D motion trajectories that connect them in time. 

\section{Related work}
\label{sec:related}

\noindent \textbf{2D and 3D motion data.} 
There has been tremendous progress on 
motion estimation from images and videos, and in particular for 2D image-space correspondence estimation. 
Most state-of-the-art methods use neural networks trained on ground truth data to predict these correspondences directly from images. While these approaches require large training datasets, synthetic data from graphics engines~\cite{dosovitskiy2015flownet, mayer2016large, harley2022particle, butler2012naturalistic, sun2021autoflow, greff2021kubric, zheng2023point}  has proven surprisingly effective at generalizing to real-world data, likely because the core task, low-level textural correspondence, is similar between the two domains.

However, the same cannot be said for 3D motion estimation, since predicting both 3D geometry and motion is often more ambiguous and can require specific prior knowledge about the real world and how it moves. 
To help address this domain gap, a number of real-world datasets have been proposed. The KITTI~\cite{geiger2013vision} and Waymo~\cite{sun2020scalability} datasets include real-world autonomous driving sequences with stereo and motion annotations derived from LiDAR and odometry information, but focus on the relatively closed domain of street scenes. Our data depicts more diverse in-the-wild scenarios. 
A number of annotated smaller-scale datasets, such as TAPVid~\cite{doersch2022tap}, TAPVid3D~\cite{koppula2024tapvid3d}, and Dycheck~\cite{gao2022monocular}, have been proposed, primarily serving as evaluation datasets for benchmarking depth estimation, 3D reconstruction, and 3D motion estimation approaches.
WSVD~\cite{wang2019web} and NVDS~\cite{NVDS} are stereo video datasets that include 
disparity maps derived from optical flow. 
While their source content is similar, our method provides richer 3D annotations beyond time-independent disparity maps, such as camera parameters and long-term 3D motion tracks.

\bfpar{Static and dynamic scene reconstruction}
The problem of reconstructing a static 3D scene has been studied for decades. Traditional 3D reconstruction methods tackle this problem by first estimating camera parameters via Structure-from-Motion~(SfM)~\cite{snavely2006photo, pollefeys2008detailed, pollefeys2004visual, agarwal2011building, schonberger2016structure, sweeney2019structure, holynski2020reducing} or SLAM~\cite{engel2017direct, campos2021orb, murartal2015orbslam, davison2007monoslam}. 
Dense scene geometry can then be estimated through Multi-view Stereo (MVS)~\cite{campbell2008using, jancosek2011multi, furukawa2010towards, furukawa2009accurate, galliani2015massively, schonberger2016pixelwise, yao2018mvsnet, yao2019recurrent} followed by surface reconstruction algorithms~\cite{hoppe1992surface, curless1996volumetric, kazhdan2006poisson}. More recently, deep neural network-based approaches have shown promising results in improving camera localization accuracy or scene reconstruction through intermediate representations such as depth maps~\cite{bloesch2018codeslam, tang2018ba, teed2024deep, teed2021droid, shen2023dytanvo, li2024megasam}, point tracks~\cite{wang2023vggsfm,lindenberger2021pixsfm}, radiance fields~\cite{lin2021barf, Fu_2024_CVPR, park2023camp, gao2024cat3d, shih2024extranerf, weber2024nerfiller}, or 3D scene coordinates~\cite{brachmann2023ace, brachmann2024acezero, leroy2024grounding, wang2024dust3r, zhang2024monst3r}. However, these methods assume the input images to be observations of a static environment, and therefore produce inaccurate geometry and camera poses for dynamic scenes.

Reconstructing dynamic scenes is more challenging since scene and object motions violate the multi-view constraints used to reconstruct static scenes. As a result, many prior works require RGBD input~\cite{bozic2020deepdeform,  newcombe2015dynamicfusion} or only recover sparse geometry~\cite{park20103d, vo2016spatiotemporal, simon2016kronecker}. Recent works tackle this problem from monocular input through video depth maps~\cite{zhang2021consistent, kopf2021rcvd, zhang2022structure}, time-varying radiance fields~\cite{park2021nerfies,park2021hypernerf,li2021neural, liu2023robust, li2023dynibar, gao2022monocular, lei2024mosca, wang2024shape}, or generative priors~\cite{wu2024cat4d}.

\bfpar{Monocular and stereo depth.}
Recent works on single-view depth prediction have shown strong zero-shot generalization to in-the-wild domains by training deep neural networks on diverse RGBD datasets~\cite{li2018megadepth, li2019learning, ranftl2021vision, yin2021learning, ranftl2020towards, ke2024repurposing, depthanything, yang2024depth, yin2023metric3d, piccinelli2024unidepth}. However, producing \textit{temporally consistent} and \textit{metric} depth from video is still challenging. To tackle this, recent works use test-time optimization~\cite{luo2020consistent, zhang2022structure} or end-to-end learning with temporal attention~\cite{kopf2021rcvd, hu2024depthcrafter, shao2024learning,NVDS}.
On the other hand, stereo images or videos are also popular input modalities for obtaining reliable metric depth maps, and various stereo matching algorithms have been proposed~\cite{birchfield1999depth, hirschmuller2002real, van2002hierarchical, klaus2006segment, sun2003stereo, pang2017cascade, chang2018pyramid, kendall2017end, zhang2019ga, li2023temporally, zhang2023temporalstereo, karaev2023dynamicstereo, jing2024matchstereovideos, wang2025sea}.
Building on these advancements, our method bridges ideas from monocular video depth estimation and stereo video processing. We use a light-weight optimization step and extend them to stereo inputs for more consistent motion estimation in metric space.

\section{Creating a dataset of 4D scenes}
\label{sec:data}

\begin{figure}
    \centering
    \includegraphics[width=\linewidth]{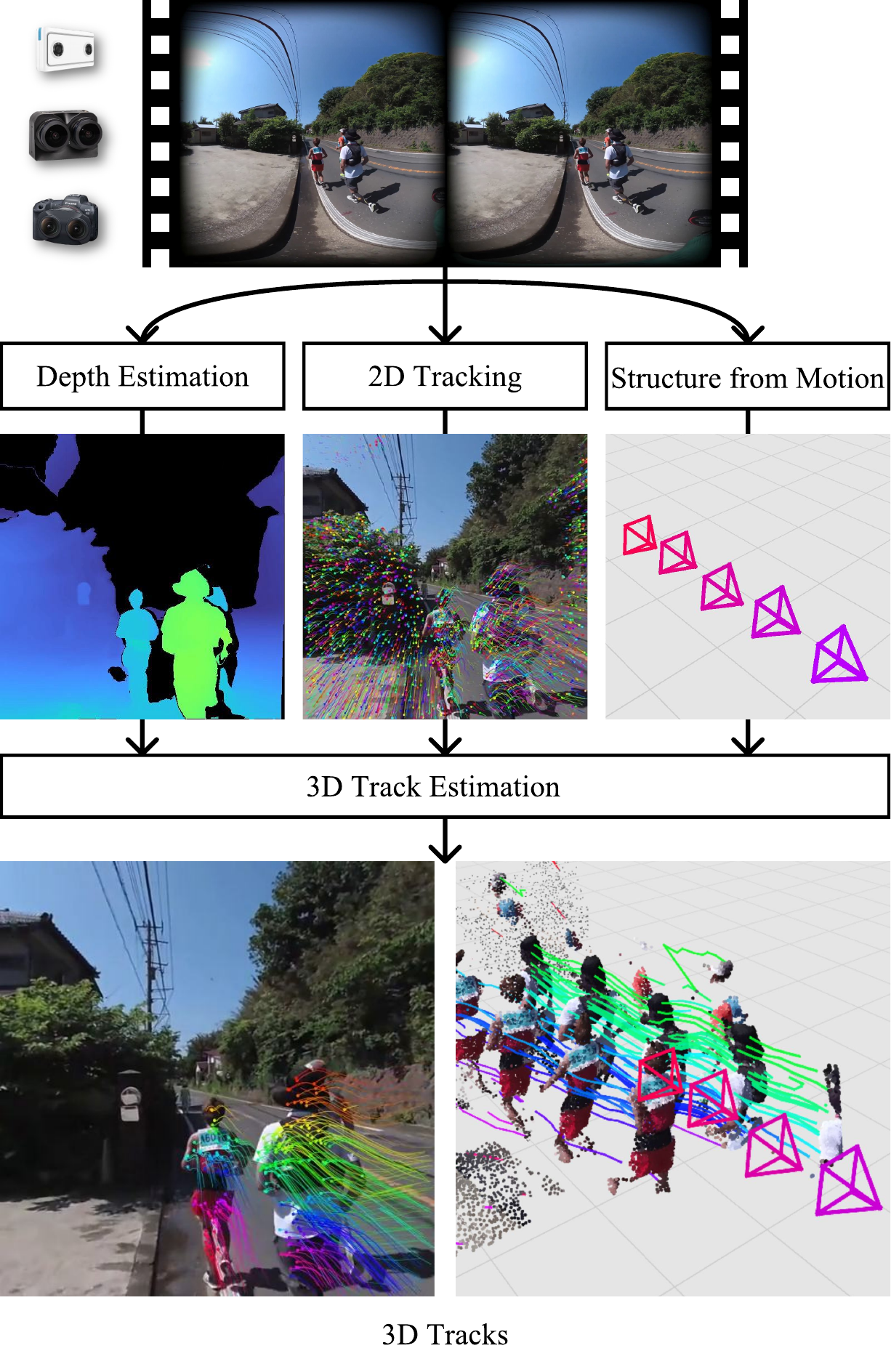}
    \vspace{-2em}
    \caption{\textbf{Data processing pipeline.} Our method starts with VR180 (wide-angle, stereoscopic) videos, and estimates metric stereo depth, 2D point tracks, and camera poses. These quantities allow the tracks to be lifted to 3D where they are filtered and denoised to produce world-space, metric 3D point trajectories.}
    \label{fig:data_pipeline}
\end{figure}

A core contribution of our work is a pipeline for extracting high-quality, pseudo-metric, 3D data from online stereoscopic fisheye videos (known as VR180 videos). 
High-resolution, wide field of view VR180 videos can be found readily online.
We show that this data is ideal for deriving rich dynamic 3D information that can power models for predicting geometry and motion from imagery.

Concretely, each instance of data starts as an $N$ frame stereo video consisting of left-right image pairs $\IB_i$ and $\IB'_i$ indexed by frame index $i\in[1,N]$. We convert these stereo pairs to a dynamic 3D point cloud with $K$ points in a world-space coordinate frame, where each point, indexed by $j\in[1,K]$, has a time-varying position $\pB_i^j$. 
As part of the process of generating this dynamic point cloud, we also extract a number of auxiliary quantities: (1) per-frame camera extrinsics, (the left camera's position $\cB_i$ and orientation $\RB_i$), (2) rig calibration for the stereo video giving the position $\cB_r$ and orientation $\RB_r$ of the right camera relative to the left camera, and (3) a per-frame disparity map $\DB_i$.

\subsection{Data Processing Pipeline} 

At a high level, our pipeline for converting a stereoscopic video into a dynamic point cloud involves estimating camera poses, stereo disparity, and 2D tracks, fusing these quantities into a consistent 3D coordinate frame, and performing several filtering operations to ensure temporal consistent, high-quality reconstructions (\Fig{data_pipeline}). In this section, we describe 
the key components of this process. %

\medskip
\noindent \textbf{SfM.} 
We start by processing the sequence of stereo frames $\IB_i \leftrightarrow \IB_i'$ to produce camera pose estimates ($\cB_i, \RB_i$). We first use a SLAM method to divide the video into shots, as in~\cite{zhou2018stereo}. For each shot, we run an incremental SfM algorithm similar to COLMAP~\cite{schonberger2016structure}. 
We initialize the stereo rig calibration $(\cB_r, \RB_r)$ to a rectified stereo pair with baseline $6.3$cm, but optimize for the calibration in bundle adjustment. 
In practice, we found that the exact stereo pair orientation can vary significantly from its nominal configuration and that optimizing the rig was critical for good results. 

\medskip 
\noindent \textbf{Depth Estimation.}
We next estimate a per-frame disparity map, operating on each frame independently. In particular, we use the estimated camera rig calibration $\cB_r, \RB_r$ to create rectified stereo pairs from the stereo fisheye video and estimate per-frame disparity~$\DB_i$ with RAFT~\cite{sun2022disentangling,
sun2021autoflow,teed2020raft}. 

\medskip
\noindent \textbf{3D Track Estimation and Optimization.}  %
We extract long-range 2D point trajectories with BootsTAP~\cite{doersch2024bootstap}. Using the camera poses $\cB_i, \RB_i$ and disparity maps $\DB_i$, we unproject these tracked points into 3D space, turning each 2D track $j$ into a 3D motion trajectory $\pB^j_1, \ldots, \pB^j_N$
 across all frames. In general, each point will usually only be tracked in a subset of frames, but for simplicity, we describe the formulation as if all points are always visible. Moreover, since subsequent steps are done independently per-track, we drop the superscript $j$ in future references. 

Since stereo depth estimation is performed per-frame, the initial disparity estimates (and therefore, the 3D track positions) are likely to exhibit high-frequency temporal jitter. 
To compensate for potentially inconsistent disparity estimates, we formulate an optimization strategy that solves for a per-frame scalar offset $\delta_i \in \mathbb{R}$ that moves each point $\pB_i$ along the ray from camera location $\cB_i$ to $\pB_i$ at frame $i$. 
This ray is denoted $\rB_i = (\pB_i - \cB_i) / ||\pB_i - \cB_i||$, and we refer to the updated location as $\pB_i' = \pB_i + \delta_i \rB_i$. 

To ensure static points remain stationary while moving tracks maintain realistic, smooth motion, avoiding abrupt depth changes frame by frame, we design an optimization objective comprising three terms: 
a static loss $\mathcal{L}_{\mathsf{static}}$, 
a dynamic loss $\mathcal{L}_{\mathsf{dynamic}}$, 
and a regularization loss $\mathcal{L}_{\mathsf{reg}}$. 
The static loss $\mathcal{L}_{\mathsf{static}}$ minimizes jitter by encouraging points to remain close to each other in world space:
\begin{equation}
\mathcal{L}_{\mathsf{static}} = \sum_{i=1}^{N} \sum_{j=1}^{N} \frac{\| \mathbf{p}_i' - \mathbf{p}_j' \|^2}{N_p'^2}
\label{eq:objective_function}
\end{equation}
where $N_p' = \sum_{i=1}^N{\|\mathbf{p}'_i\|} / N$ is a normalizing factor.
The dynamic loss term reduces jitter by minimizing acceleration along the camera ray through a discrete Laplacian operator:
\begin{equation}
\mathcal{L}_{\mathsf{dynamic}} = \sum_{i=1}^{N} \sum_{\Delta\in\mathcal{W}} \left[ \left( \mathbf{p}_{i+\Delta}' - 2\mathbf{p}_i' + \mathbf{p}_{i-\Delta}' \right)^\top \mathbf{r}_i \right]^2
\label{eq:dynamic_objective}
\end{equation}
where the acceleration along the ray is calculated over
multiple window sizes $\mathcal{W}=\{1,3,5\}$.

The two loss terms are weighted by a track-dependent function, $\sigma(m)$, where $m$ is a measure of the motion magnitude of the track. 
Motion is measured in 2D rather than 3D because distant points can appear to have a larger 3D motion due to noise amplification at low disparities.
Specifically, we project the 3D motion trajectory between time $i - w_o$ and the current time $i$ into 2D image-space at time $i$, %
and calculate the track's motion magnitude $m$ as the 90$^{th}$ percentile of the track's trail length across all frames. The track trail length for a frame is measured by projected 3D points along the track to the current frame as if the camera is \emph{static} in a window of $w_o=16$ frames, 
\begin{equation}
\label{eqn:trail_length_def}
    m = \mathsf{Percentile}_{i=1:N}^{90}\left[\max_{w=1:w_o}\|\pi_i(\pB_i) - \pi_i(\pB_{i-w})\|\right]
\end{equation}
where $\pi_i(\cdot)\in \mathbb{R}^2$ gives the projected pixel location of a 3D point on camera $\cB_i$'s image plane.
The weighting function $\sigma(m)$ is defined as $\sigma(m) = \frac{1}{1 + \exp(m - m_0)}$ where $m_0 = 20$. 
Finally, to encourage faithfulness to the originally estimated disparities, we regularize the displacements in disparity space:
\begin{equation}
    \mathcal{L}_{\mathsf{reg}} = \lambda_{\mathsf{reg}} \sum_{i=1}^{T} \left( \frac{1}{\delta_i + \|\mathbf{p}_i-\mathbf{c}_i\|} - \frac{1}{\|\mathbf{p}_i-\mathbf{c}_i\|} \right)^2,
\end{equation}
where use of disparity space reflects the fact that the measurements themselves originate as disparities. Practically, the impact of the use of disparity is that larger deviations are tolerated at more distant points, where depth is intrinsically more uncertain.

\begin{figure}[t]
    \centering
    \includegraphics[width=\linewidth]{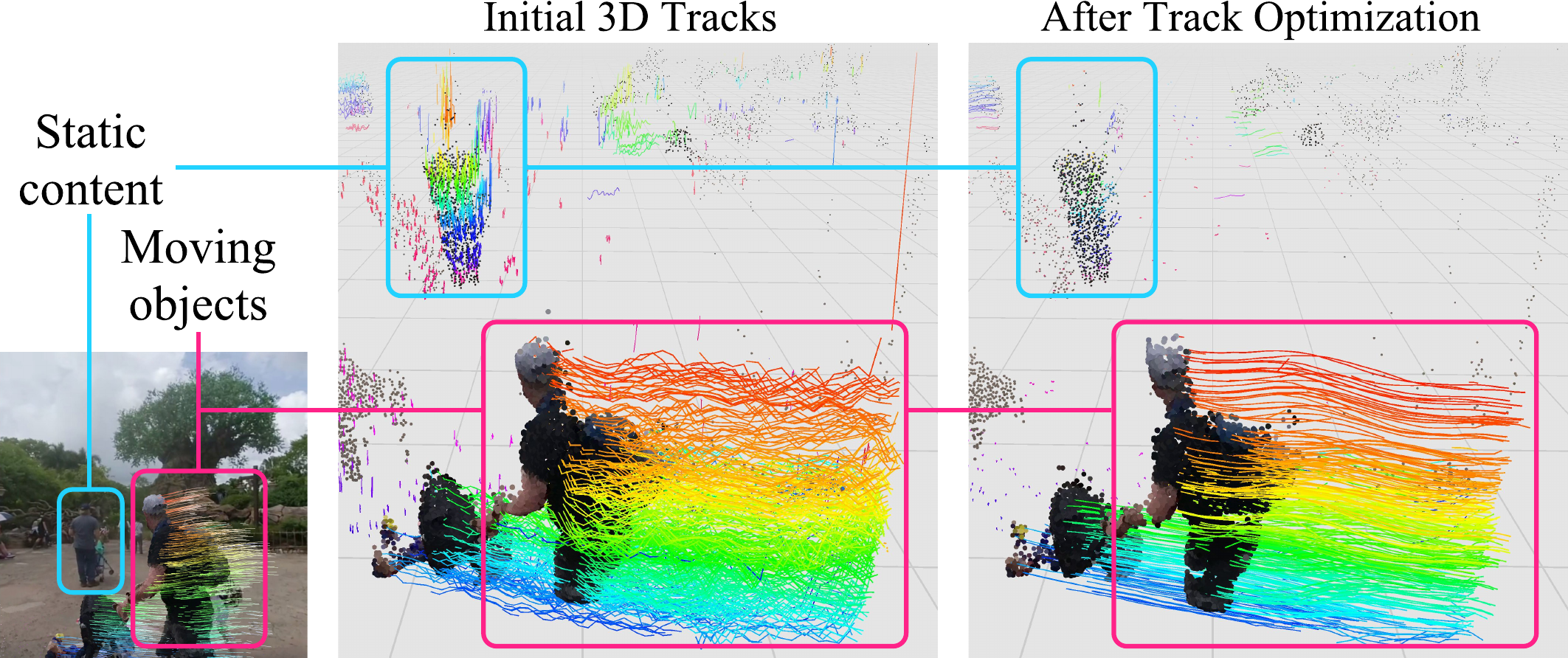}
    \caption{\textbf{Effect of track optimization.} Comparing motion trajectories before and after track optimization, we see that optimization resolves the high-frequency jitter along camera rays, affecting both static and dynamic content. After optimization, static content has static tracks, and dynamic tracks are less noisy.}
    \label{fig:track_optimization}
\end{figure}

The full objective function is
\begin{equation}
\min_{\{\delta_i\}_{i=1}^N} \sigma(m)\mathcal{L}_{\mathsf{static}} + (1-\sigma(m))\mathcal{L}_{\mathsf{dynamic}} + \mathcal{L}_{\mathsf{reg}}.
\label{eqn:objective}
\end{equation}
We set $\lambda_{\mathsf{reg}}=10^{-4}$ and optimize Eqn.~\ref{eqn:objective} using Adam with a learning rate of 0.05 for 100 steps.
The effect of track optimization is shown in \Fig{track_optimization}. The optimized motion is smoother and does not contain high frequency noise.

\begin{figure*}[t]
\vspace{-1em}
    \centering
    \includegraphics[width=\textwidth]{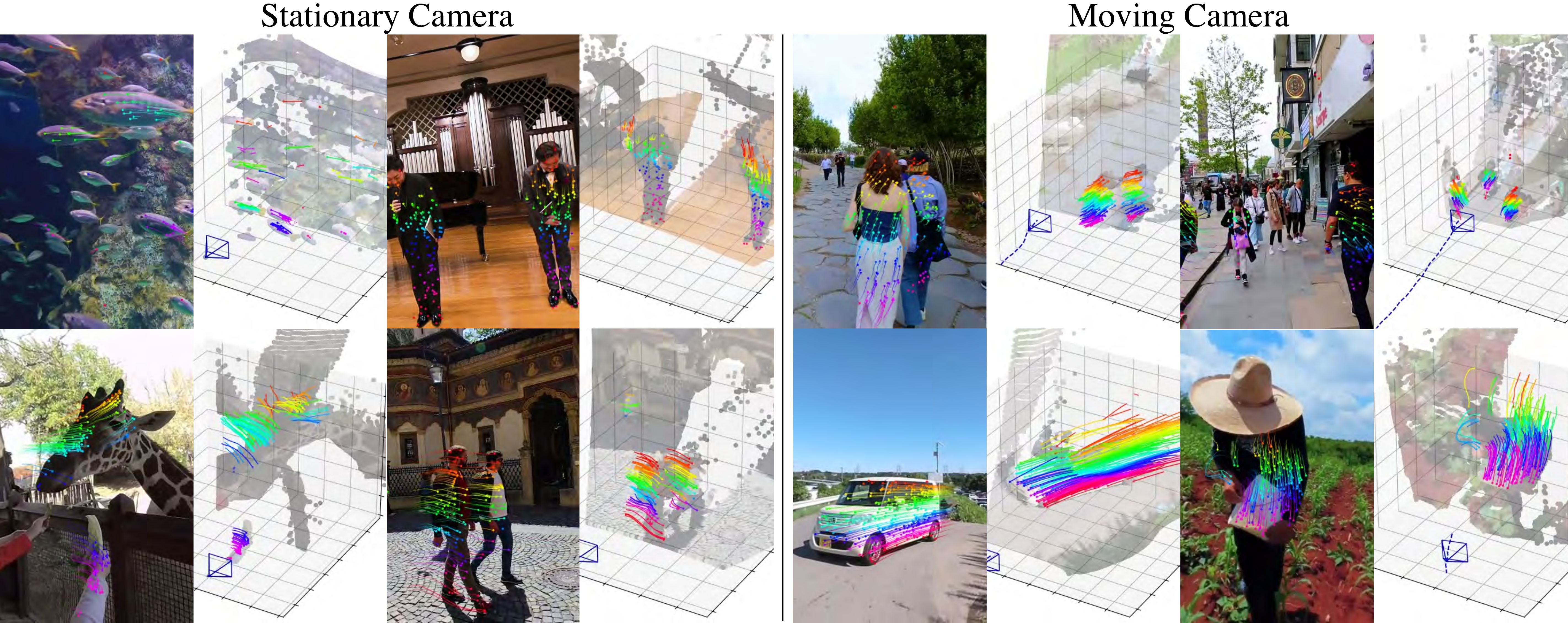}
    \caption{\textbf{Diverse motion:} \dataset captures a wide variety of types of moving objects: swimming fish, walking pedestrians, moving vehicles, a farmer sowing seeds, etc. It includes source videos captured with both stationary (left) and moving (right) cameras.}
    \label{fig:motion_distribution}
\end{figure*}
\medskip
\noindent \textbf{Implementation details.} %
{\it Shot-selection.} Rather than work with the full video, we break the footage into discrete, trackable shots using ORB-SLAM2's stereo estimation mode~\cite{murartal2015orbslam} following~\cite{zhou2018stereo}. 
{\it Field of view.} While estimating pose, we use a $140^\circ$ FoV fisheye format, which we found to capture more of the (usually static) background and less of the (often dynamic) foreground, yielding more reliable camera poses. 
{\it Stereo confidence checks.} We discard pixels where the $y$-component of RAFT flow is more than 1 pixel (since rectified stereo pairs should have perfectly horizontal motion) and where the stereo cycle consistency error is more than 1 pixel (since such pixels are unreliable). 
{\it Dense 2D tracks.} To obtain dense tracks, we run BootsTAP with dense query points: for every 10th frame, we uniformly initialize $128\times128$ query points on frames of resolution 512 $\times$ 512. We then prune redundant tracks that overlap on the same pixel.
{\it Drifting tracks.} Since 2D tracks can drift on textureless regions, we discard moving 3D tracks that correspond to certain semantic categories (\textit{e.g.}, ``walls'', ``building'', ``road'', ``earth'', ``sidewalk''), detected by DeepLabv3~\cite{chen2017rethinking} on ADE20K classes~\cite{zhou2017scene, zhou2019semantic}.

\medskip
\noindent \textbf{Filtering details.} A fraction of the video clips that are processed are unsuitable because they either 
(1) are not videos, but entirely static images, 
(2) contain cross-fades, or 
(3) have text or other synthetic graphics. To discard text and title sequences, we avoid creating video clips from the start and ends of the source videos. We identify cross-fades by running SIFT~\cite{lowe2004sift} matching through the video at multiple temporal scales and discarding video clips with static camera but with fewer than 5 SIFT matches between frames that are 5 seconds apart.

\begin{figure}[t]
    \centering
    \includegraphics[width=\linewidth]{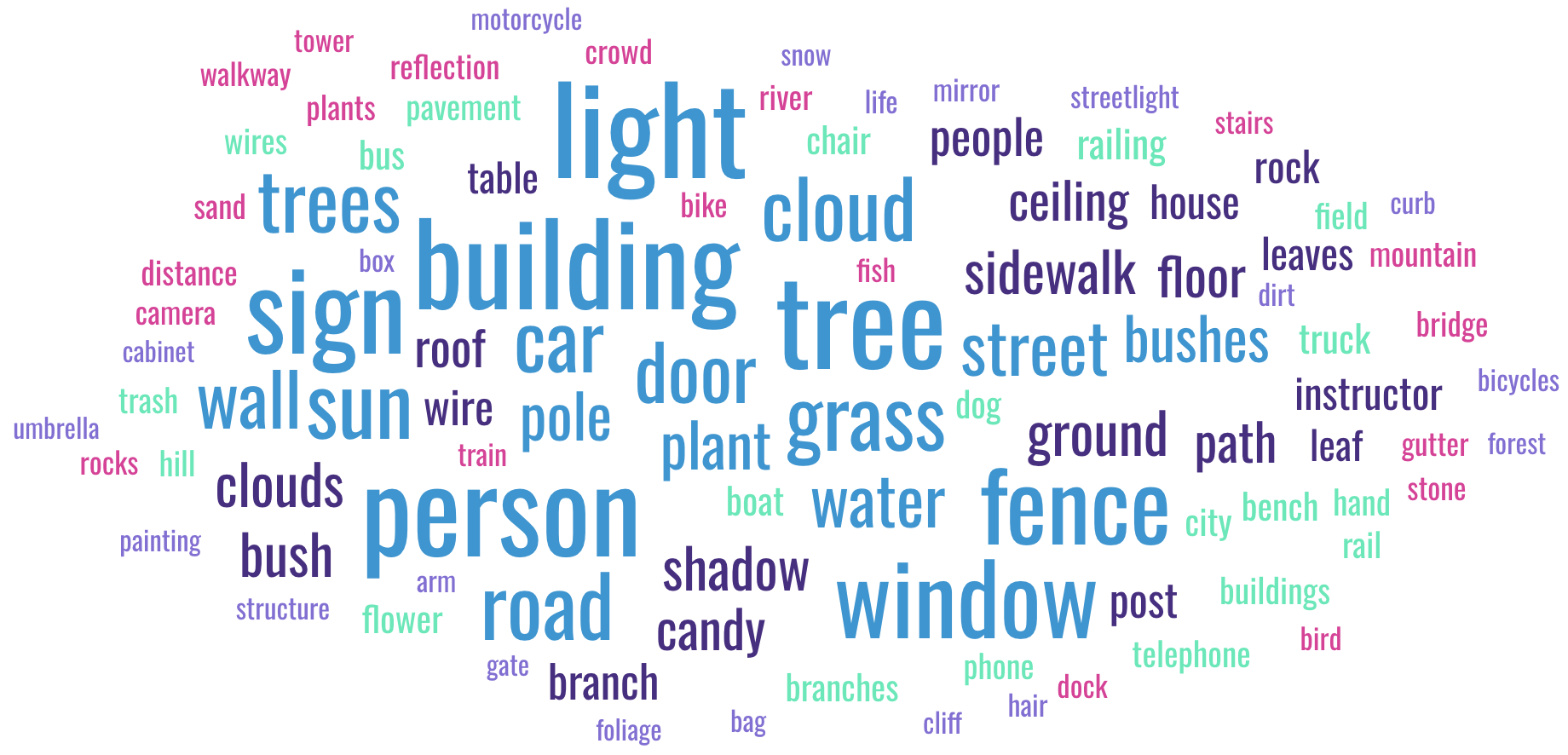}
    \caption{\textbf{Diverse scene content:} A word cloud of captioned frames from our dataset shows our data is diverse, including a variety of common objects seen in videos.}
    \label{fig:wordcloud}
\end{figure}

\subsection{Stereo4D Dataset}

\Fig{motion_distribution} illustrates a subset of videos and reconstructions from a dataset processed with the above pipeline, encompassing more than 100K clips capturing everyday scenes and activities. To visualize the range of content, we used an automatic captioning system to generate captions for the dataset and created a word cloud (\Fig{wordcloud}) highlighting the most frequently observed objects.
See supplementary for more statistics and ablations on track optimization. 

\section{Learning a prior on how things move}
We now describe our method for predicting dynamic 3D point clouds from pairs of images, and how we train it with our Stereo4D data.
Our model is based on \duster~\cite{wang2024dust3r}, which predicts a 3D point cloud for a static scene from images. Given two input images, it uses a ViT-based architecture~\cite{dosovitskiy2020image} to extract image features and uses a transformer-based decoder to cross-attend features from two images, and then use a downstream \textit{pointmap} decoder to output pointmaps for the two images, aligned in the first image's coordinate frame. 

\begin{figure}[t]
    \centering
    \includegraphics[width=\linewidth]{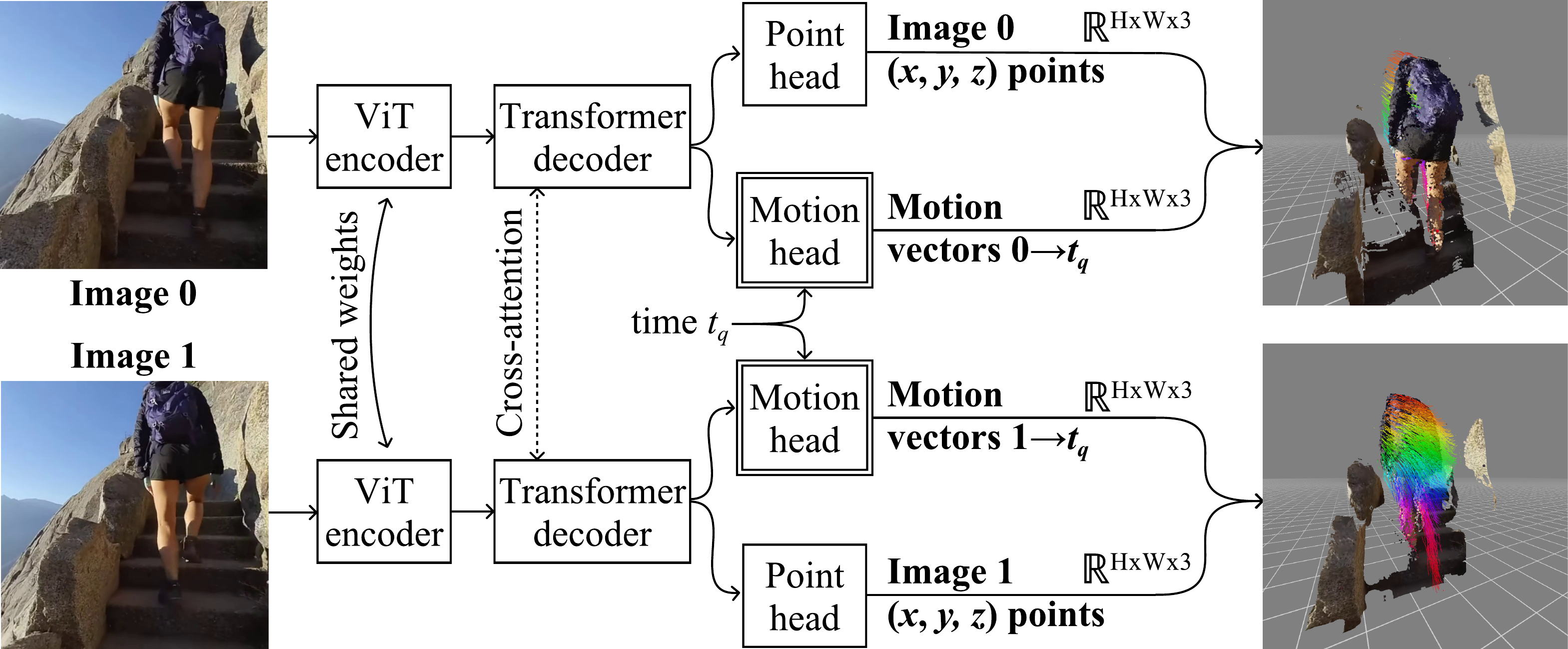}
    \caption{\textbf{\method architecture.} Given two images $(\mathbf{I}_0, \mathbf{I}_1)$ of a dynamic scene and a desired target time $t_q$, the images are passed through a ViT encoder and transformer decoder. The resulting features are processed by (1) a pointmap head that predicts 3D points in the coordinate frame of $\mathbf{I}_0$, and (2) a 3D motion head that predicts the motion of all points to the target time $t_q$. A double outline indicates a new component compared to DUSt3R.}
    \label{fig:method}
\end{figure}

\bfpar{\method model.}
While \duster focuses on static scene structure, our proposed \method method, illustrated in \Fig{method}, works with dynamic scenes by adding a \textit{motion} head that predicts how the points move between two frames. 
As input, \method accepts two images: $\mathbf{I}_0$ at time $t_0$, and $\mathbf{I}_1$ at time $t_1$ (where $t_0$ and $t_1$ may be seconds apart). 
It also accepts an intermediate query time $t_q \in [0,1]$; the motion head is asked to predict 3D scene flow from the two input frames to query time $t_q$, as described below.

Like \duster, \method begins by encoding the images with a shared ViT and cross-attention decoder, producing global features $G^0$ and $G^1$ for $\mathbf{I}_0$ and $\mathbf{I}_1$, respectively.
Each feature embedding can be converted into geometry using \duster's point head: e.g., for image $\mathbf{I}_0$, the point head produces a pointmap $\PB^0 \in \mathbb{R}^{H \times W \times 3}$ representing the geometry at time $t_0$, as well as a point confidence map $\CB^0_\mathsf{point} \in \mathbb{R}^{H \times W}$. 
Each point cloud is predicted in the coordinate frame of $\mathbf{I}_0$, but \emph{at the time of its respective image} (so, the two point clouds may differ due to scene motion).

We add a separate \textit{motion} head in parallel to the original point head, to predict a map of 3D displacement vectors (that is, a scene flow map, which we refer to as a \emph{motion map}) for each pointmap. 
The motion map should displace each input frame to an intermediate time $t_q \in [0,1]$, where $t_q = 0$ corresponds to $t_0$, the time of $\mathbf{I}_0$, and similarly for $t_q = 1$, $t_1$, and $\mathbf{I}_1$. 
The motivation for predicting motion to an intermediate time (inclusive of the endpoints) is twofold: first, it leads to a more general prediction task where we can predict a full motion trajectory between two frames, and second, it allows us to use partial ground truth 3D trajectories as supervision; not all trajectories may span all the way from $t_0$ to $t_1$, but may span through some intermediate time.

For each image $\IB_v$ (with $v \in \{0,1\}$), the network outputs a 3D motion map $\mathbf{M}^{v\to t_q}$ for the corresponding pointmap from $t_v$ to $t_q$ with corresponding motion confidence map $\CB_\mathsf{mot}^{v}\in\mathbb{R}^{H\times W}$. This prediction is based on the global feature $G^v$ as well as an embedding of the query time $\texttt{emb}(t_q)$. We use positional embedding~\cite{vaswani2017attention} to encode time $t_q$ to a 128-D vector and inject it to the motion features in the motion head via linear projection layers.

\bfpar{Training objective.} We use the same confidence-aware scale-invariant 3D regression loss
as in \duster. 
We first normalize both the predicted and ground truth pointmaps using scale factors $z=\text{norm}(\mathbf{P}^0, \mathbf{P}^1)$ and $\bar{z}=\text{norm}(\bar{\mathbf{P}}^0, \bar{\mathbf{P}}^1)$, respectively (where a bar, e.g., $\bar{\mathbf{P}}^0$, denotes a ground truth quantity, and where `$\text{norm}$' computes the average distance between a set of points and the world origin). 
We scale the motion maps with the same scales $z$ and $\bar{z}$. 
Following \duster, we compute a Euclidean distance loss on the pointmap, setting $\loss{point}$ to
{\small
\begin{equation}
    \sum_{v\in\{0,1\}}\sum_{i\in\mathcal{D}^v}\CB_{\mathsf{point},i}^v\left\|\frac{1}{z}\mathbf{P}_i^v - \frac{1}{\bar{z}}\bar{\mathbf{P}}_i^v\right\| - \alpha_p\log\CB_{\mathsf{point},i}^v
    \label{eqn:loss_point}
\end{equation}}
where $\mathcal{D}^v$ corresponds to the valid pixels where ground truth is defined and $\alpha_p$ is a weighting hyperparameter. 
We additionally compute a Euclidean distance loss on the position \emph{after motion}, which encourages the network to learn correct displacements. This loss $\loss{motion}$ is defined as
{\small
\begin{equation}\label{eqn:loss_motion}
    \sum_{v\in\{0,1\}}\sum_{i\in\mathcal{D}^v}\CB_{\mathsf{mot},i}^v\left\|\frac{1}{z}\mathbf{P}_i^{v\to t_q} - \frac{1}{\bar{z}}\bar{\mathbf{P}}_i^{v\to t_q}\right\| \\
    - \alpha_m\log\CB_{\mathsf{mot},i}^v,
\end{equation}}
where $ \mathbf{P}_i^{v\to t_q} = \mathbf{P}_i^v + \mathbf{M}_i^{v\to t_q}$.

\bfpar{Training details.}
We initialize our network with \duster weights and initialize the motion head with the same weights as the point head. We finetune for 49k iterations, with batch size 64, learning rate 2.5e-5, optimized by Adam with weight decay 0.95. During training, we randomly sample pairs of video frames that are at most 60 frames apart. 
The weight for the confidence loss in Eqn~\ref{eqn:loss_point}-\ref{eqn:loss_motion} is $\alpha_m = \alpha_p = 0.2$. The model is trained on tracks extracted from both 60$^\circ$ FoV videos for (higher quality) and 120$^\circ$ FoV videos for (larger coverage). 

\section{Experiments}

We conduct a series of experiments to validate the effectiveness of our proposed data and techniques.
First, we evaluate 
our proposed real-world \dataset data mined from VR180 videos on the \method task.
In particular, we compare models that are individually trained with our real-world data and with synthetic data, and we show that our data enables model learning more accurate 3D motion priors (\Sec{motion_eval}). 
Second, we show that our trained model that adapts \duster has strong generalization to in-the-wild images of dynamic scenes, and enables accurate predictions of underlying geometry (\Sec{structure_eval}). See supplementary for more ablations on \method design.

\begin{figure*}[ht]
\vspace{-1em}
    \centering
    \includegraphics[width=\linewidth]{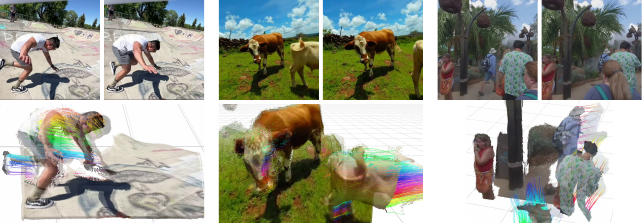}
    \caption{\textbf{Testing on held out examples from \dataset.} We visualize image pairs and corresponding dynamic 3D point clouds predicted by \method. 
    It recovers accurate 3D shape and complex scene motion for objects such as people breakdancing and cows walking.}
    \label{fig:result-wall-stereo4dtest}
\end{figure*}

\subsection{3D motion prediction} \label{sec:motion_eval}

\noindent \textbf{Baselines and metrics.} 
To evaluate the efficacy of our data paradigm on motion prediction, we primarily compare \method trained on \dataset to the same network trained on a synthetic dataset, PointOdyssey~\cite{zheng2023point}. 
PointOdyssey contains ground truth depth maps and 3D motion tracks rendered from an animation engine;
we supervise the model with this data using the same hyperparameter settings as described above. 
During inference, given two video frames sampled from a video of a dynamic scene, we compare 3D end-point-error (EPE) between ground truth and predicted 3D motion vectors. 
We also 
compute the fraction of 3D points that have motion within 5 cm and 10 cm compared to ground truth ($\delta_{3D}^{0.05}, \delta_{3D}^{0.10}$), following~\cite{teed2021raft3d,wang2024shape}. 
Since our model outputs point clouds up to an unknown scale, we align each prediction with the ground truth through a median scale estimate.
We evaluate models trained on each of these two data sources on a held-out \dataset test-set, as well as on Arial Digital Twin (ADT)~\cite{pan2023aria} data containing scene motion, processed by the TapVid3D benchmark~\cite{koppula2024tapvid3d}. 
As test data, we randomly sample pairs of frames that are at most 30 frames apart from both \dataset and ADT.

\begin{table}[!t]
\centering
\footnotesize
\renewcommand{\arraystretch}{0.95}
\renewcommand{\tabcolsep}{2.pt}

\resizebox{\linewidth}{!}{
\begin{tabular}{@{}lcccccc@{}}
\toprule
  & \multicolumn{3}{c}{Stereo4D} & \multicolumn{3}{c}{ADT} \\ 
 
\cmidrule(lr){2-4} \cmidrule(lr){5-7}
 Method & $\text{EPE}_\text{3D}$$\downarrow$&	$\delta_\text{3D}^{0.05}$$\uparrow$ &$\delta_\text{3D}^{0.10}$$\uparrow$& $\text{EPE}_\text{3D}$$\downarrow$&	$\delta_\text{3D}^{0.05}$$\uparrow$ &$\delta_\text{3D}^{0.10}$$\uparrow$  \\ 
\midrule

  \method (PointOdyssey) &0.6191 & 11.61 & 20.25 & 0.3126 & 8.56 & 18.03\\ 
  \method (\dataset) & \textbf{0.1110} & \textbf{65.07 }& \textbf{75.18} & \textbf{0.1231 }& \textbf{51.98} &\textbf{65.20} \\ 
\bottomrule
\end{tabular}
}
\caption{{\bf Synthetic vs.\ Real Training Data.} Compared to synthetic data (PointOdyssey~\cite{zheng2023point}), training on \dataset improves \method's ability to generalize to real-world motion.}
\label{tab:motion_3d_eval}
\end{table}

\begin{figure}[ht]
    \centering
    \includegraphics[width=\linewidth]{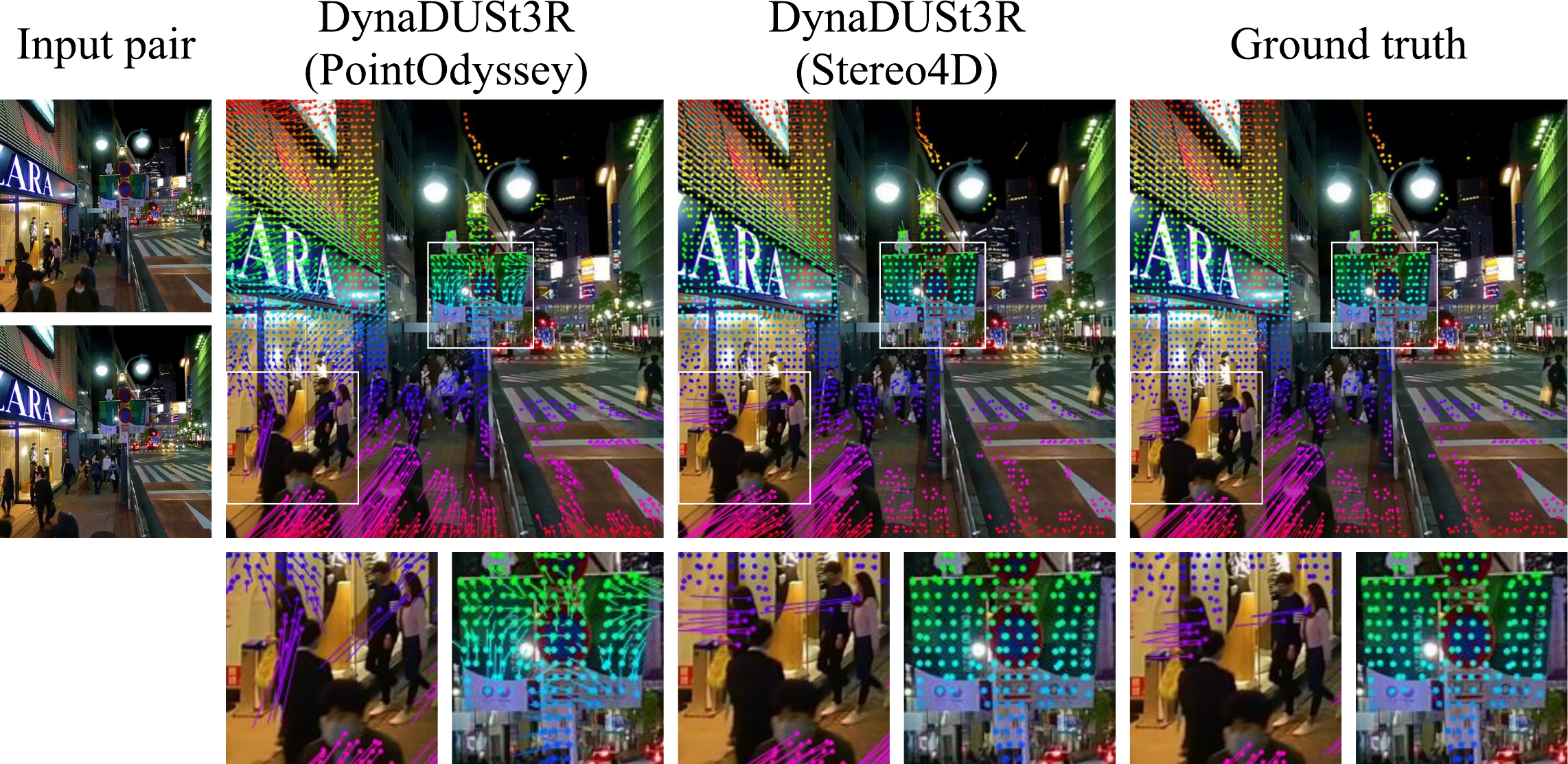}
    \caption{\textbf{Qualitative comparisons, 3D motion on the \dataset.} We compare variants of \method trained on different data sources. The PointOdyssey-trained model incorrectly predicts significant 3D motion on static elements such as the building wall and the banners near the streetlight, while the \dataset-trained model correctly predicts these elements as stationary. The \dataset model also makes more precise motion predictions for dynamic objects, such as humans with large movements (bottom row).}
    \label{fig:compare-stereo4d}
\end{figure}

\bfpar{Quantitative results.}
We show numerical results for two-frame 3D motion prediction 
in \Tab{motion_3d_eval}. 
\method trained on real-world data achieves better generalization and outperforms the baseline trained on PointOdyssey significantly across all 
metrics. This suggests the potential of our data for more effective learning of real-world 3D motion priors.

\begin{figure}[ht]
    \centering
    \includegraphics[width=\linewidth]{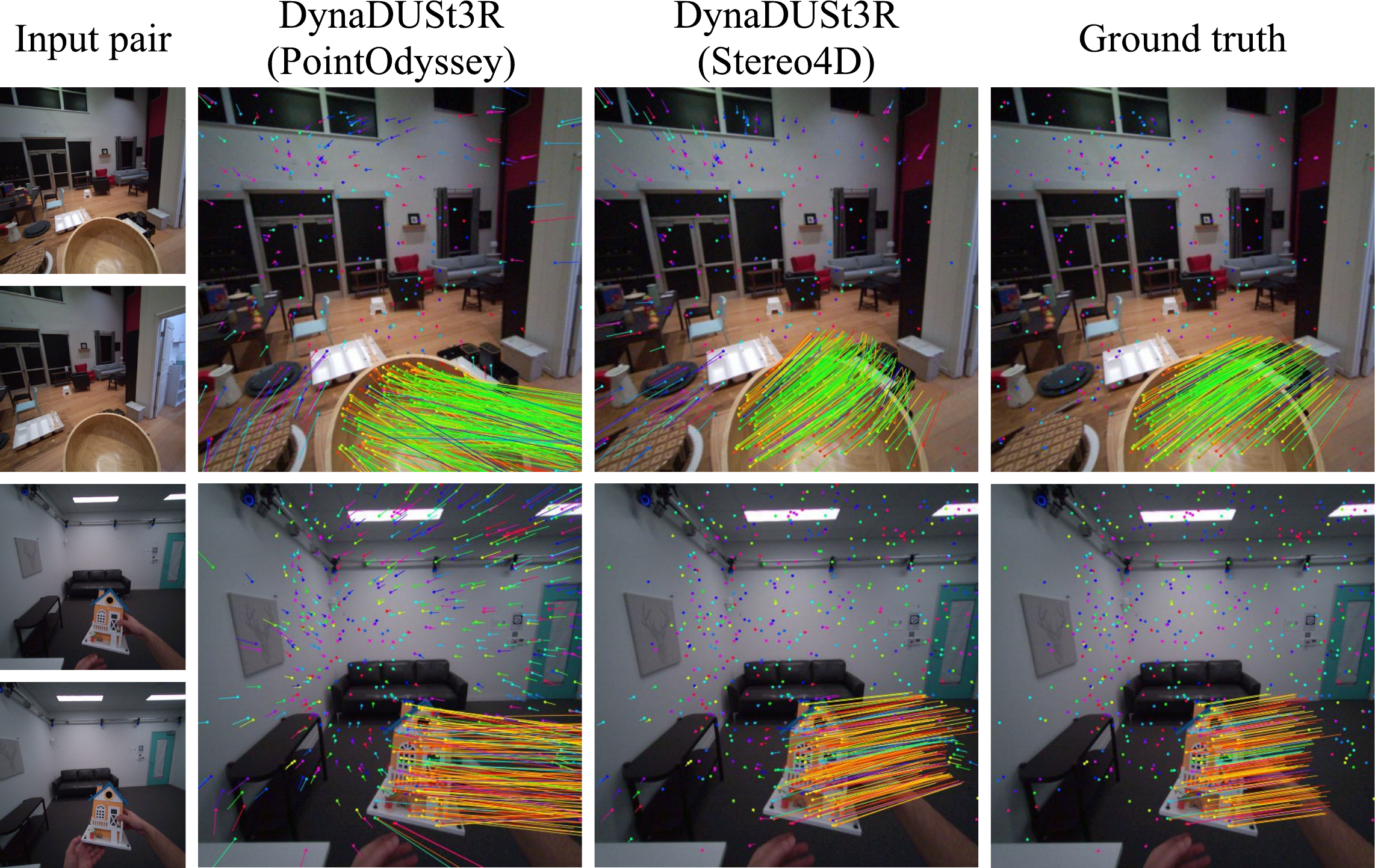}
    \caption{\textbf{Qualitative comparisons of predicted 3D motion on ADT~\cite{pan2023aria}.} \method trained on \dataset produces more accurate 3D motion compared to training on PointOdyssey.} %
    \label{fig:compare-stereo4d-adt}
\end{figure}

\bfpar{Qualitative results.}
\Fig{result-wall-stereo4dtest} shows example results for three dynamic scenes in our \dataset test set, including visualizations of 3D point clouds and motion tracks.
\method produces accurate 3D shape and motion tracks over the timespan defined by the two input images. Despite the inputs being two sparse images, our architecture enables querying intermediate motion states, resulting in continuous and potentially non-linear motion trajectories.%

We also qualitatively compare predicted 3D motion tracks between \method networks trained on \dataset and on PointOdyssey, by projecting their predicted 3D motion vectors into 2D image space.
\Fig{compare-stereo4d} and \Fig{compare-stereo4d-adt} show comparisons on the \dataset and ADT test set respectively. \method trained on \dataset produces more accurate 3D motion estimates for both static and moving objects. For instance, \method trained on PointOdyssey produces non-zero motion for the stationary street banner and erroneous motions for the walking people in \Fig{compare-stereo4d}.

\begin{figure*}[ht]
    \centering
    \includegraphics[width=\linewidth]{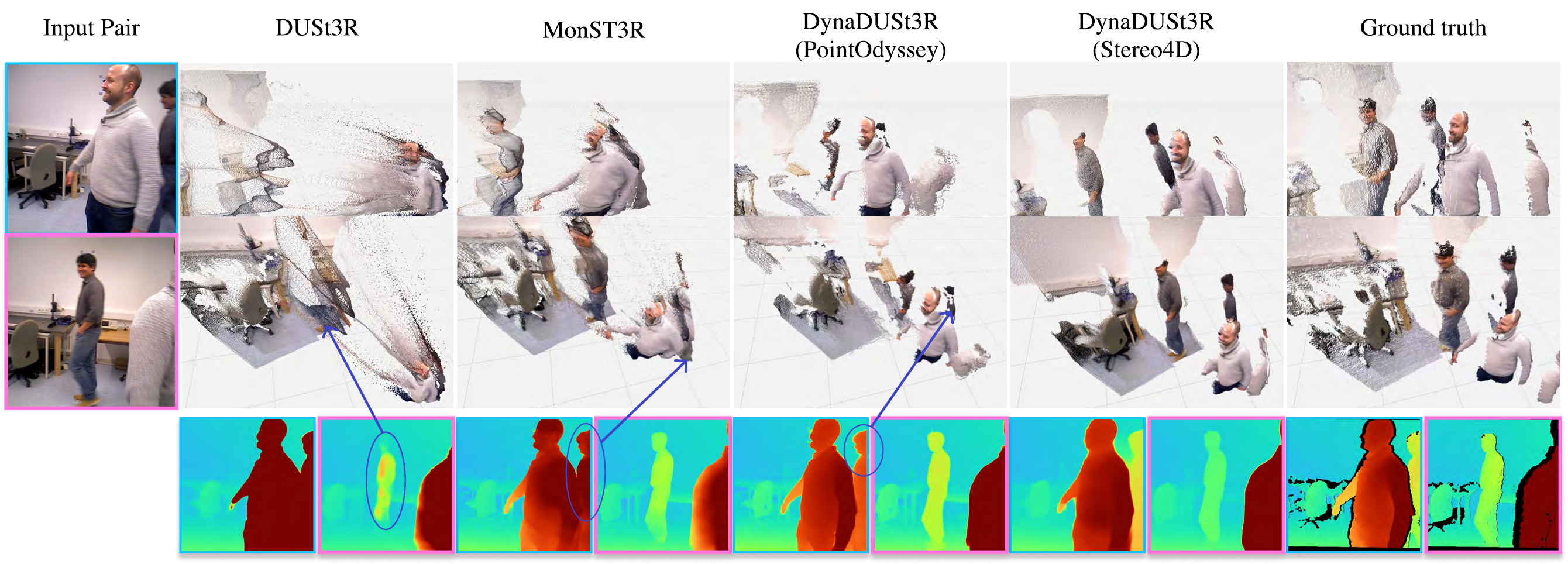}
    \caption{\textbf{Qualitative comparison, 3D structure on Bonn~\cite{palazzolo2019iros}.}  From left to right, we show an input image pair, predictions from different methods, and the ground truth geometry. 
    The top two rows show 3D point clouds from two viewpoints, where we show the union of the pointmaps for the two input time steps. The bottom row shows the corresponding disparity for two input images. 
    Compared to all the other methods, \method trained on \dataset achieves better 3D structure predictions with finer details.}
    \label{fig:compare-bonn}
\end{figure*}

\subsection{Structure prediction} \label{sec:structure_eval}
\noindent \textbf{Baseline and metrics.}
We evaluate the quality of predicted 3D structure by comparing depth maps predicted by \duster~\cite{wang2024dust3r}, \monster~\cite{zhang2024monst3r}, and \method trained on \dataset or PointOdyssey. 
\duster is designed to predict aligned point clouds from two input images of a static scene.
\monster, a concurrent approach, extends \duster to handle dynamic scenes by predicting time-varying point clouds without modeling motion.

We evaluate predicted depth accuracy on the Bonn~\cite{palazzolo2019iros} dataset and our held-out test set, where we sample two views that are 30 frames apart from a video. 
Since we focus on the two-frame case, we do not apply any post-optimization to the network outputs.
In addition, since all methods predict 3D point clouds in the coordinate frame of the first image, we include the two points clouds predicted from both input frames in our evaluation. We use standard depth metrics, including absolute relative error (Abs Rel) and percentage of inlier points $\delta<1.25$, following prior work~\cite{NVDS, zhang2024monst3r}.  We use the same median alignment as before to align the predicted depth map with the ground truth.

\bfpar{Quantitative comparisons.}
We show quantitative comparisons of depth predicted by different methods in \Tab{depth_eval}. \method trained on \dataset outperforms all other baselines by a large margin. In particular, we demonstrate improved depth prediction on the unseen Bonn dataset.

\bfpar{Qualitative comparisons.}
We provide additional visual comparisons in \Fig{compare-bonn}, where we visualize ground truth 3D point clouds and predictions from our approach and the other three baselines at different input time steps. \duster predicts inaccurate depth relationships for the two moving people, while \monster and \method trained on PointOdyssey both predict distorted scene geometry. In contrast, our model trained on \dataset produces 3D structure that most closely resembles the ground truth.

\begin{table}[!t]
\centering
\footnotesize
\renewcommand{\arraystretch}{0.95}
\renewcommand{\tabcolsep}{2.pt}

\resizebox{\linewidth}{!}{
\begin{tabular}{@{}lcccc@{}}
\toprule
  & \multicolumn{2}{c}{Stereo4D} & \multicolumn{2}{c}{Bonn~\cite{palazzolo2019iros}}  \\ 
 
\cmidrule(lr){2-3} \cmidrule(lr){4-5}
 Method & Abs Rel$\downarrow$ &$\delta<1.25$$\uparrow$ &Abs Rel$\downarrow$ &$\delta<1.25$$\uparrow$  \\ 
\midrule
\duster~\cite{wang2024dust3r} &0.2696 & 67.77 & 0.1098& 84.93 \\
\monster~\cite{zhang2024monst3r} &0.1939 & 72.56 & 0.0721&92.60\\
Ours (PtOdyssey) & 0.3858 & 61.87 & 0.0691 &95.94\\ 
Ours (\dataset) &\textbf{0.1032} & \textbf{87.93} & \textbf{0.0653} & \textbf{96.02} \\ 
\bottomrule
\end{tabular}
} 

\caption{{\bf Quantitative comparison, depth maps.} \method trained on our \dataset data surpasses the performance of the model trained on PointOdyssey~\cite{zheng2023point}, as well as \duster and \monster under challenging sparse view settings.
}\label{tab:depth_eval}
\end{table}

\section{Discussion and Conclusion}
\noindent \textbf{Limitations.} 
Our data curation pipeline and trained model have limitations. 
The quality of the long-range 3D motion tracks depends on the accuracy of optical flow and 2D point tracking and may degrade for distant background regions or objects occluded for long periods.
Additionally, \method is a non-generative model that only operates on two-frame inputs. 
Extending our model to video input by adopting an extra global optimization~\cite{zhang2024monst3r} or integrating generative priors for modeling ambiguous motion content is a promising future direction.

\bfpar{Conclusion.}
We presented a pipeline for mining high-quality 4D data from Internet stereoscopic videos. Our framework automatically annotates each real-world video sequence with camera parameters, 3D point clouds, and long-range 3D motion trajectories by consolidating different noisy structure and motion estimates derived from videos.  Furthermore, we show that training a variant of \duster on our real-world 4D data enables more accurate learning of 3D structure and motion in dynamic scenes, outperforming other baselines.

{
    \small
    \bibliographystyle{ieeenat_fullname}
    \bibliography{main}

\begin{thebibliography}{119}
\providecommand{\natexlab}[1]{#1}
\providecommand{\url}[1]{\texttt{#1}}
\expandafter\ifx\csname urlstyle\endcsname\relax
  \providecommand{\doi}[1]{doi: #1}\else
  \providecommand{\doi}{doi: \begingroup \urlstyle{rm}\Url}\fi

\bibitem[Achiam et~al.(2023)Achiam, Adler, Agarwal, Ahmad, Akkaya, Aleman,
  Almeida, Altenschmidt, Altman, Anadkat, et~al.]{achiam2023gpt}
Josh Achiam, Steven Adler, Sandhini Agarwal, Lama Ahmad, Ilge Akkaya,
  Florencia~Leoni Aleman, Diogo Almeida, Janko Altenschmidt, Sam Altman,
  Shyamal Anadkat, et~al.
\newblock {GPT-4} technical report.
\newblock \emph{arXiv preprint arXiv:2303.08774}, 2023.

\bibitem[Agarwal et~al.(2011)Agarwal, Furukawa, Snavely, Simon, Curless, Seitz,
  and Szeliski]{agarwal2011building}
Sameer Agarwal, Yasutaka Furukawa, Noah Snavely, Ian Simon, Brian Curless,
  Steven~M Seitz, and Richard Szeliski.
\newblock Building {Rome} in a day.
\newblock \emph{Communications of the ACM}, 2011.

\bibitem[Birchfield and Tomasi(1999)]{birchfield1999depth}
Stan Birchfield and Carlo Tomasi.
\newblock Depth discontinuities by pixel-to-pixel stereo.
\newblock \emph{IJCV}, 1999.

\bibitem[Bloesch et~al.(2018)Bloesch, Czarnowski, Clark, Leutenegger, and
  Davison]{bloesch2018codeslam}
Michael Bloesch, Jan Czarnowski, Ronald Clark, Stefan Leutenegger, and Andrew~J
  Davison.
\newblock {CodeSLAM}—learning a compact, optimisable representation for dense
  visual {SLAM}.
\newblock In \emph{IEEE Conf. Comput. Vis. Pattern Recog.}, 2018.

\bibitem[Bochkovskii et~al.(2024)Bochkovskii, Delaunoy, Germain, Santos, Zhou,
  Richter, and Koltun]{bochkovskii2024depth}
Aleksei Bochkovskii, Ama{\"e}l Delaunoy, Hugo Germain, Marcel Santos, Yichao
  Zhou, Stephan~R Richter, and Vladlen Koltun.
\newblock Depth {Pro}: Sharp monocular metric depth in less than a second.
\newblock \emph{arXiv preprint arXiv:2410.02073}, 2024.

\bibitem[Bozic et~al.(2020)Bozic, Zollhofer, Theobalt, and
  Nie{\ss}ner]{bozic2020deepdeform}
Aljaz Bozic, Michael Zollhofer, Christian Theobalt, and Matthias Nie{\ss}ner.
\newblock {DeepDeform}: Learning non-rigid {RGB-D} reconstruction with
  semi-supervised data.
\newblock In \emph{IEEE Conf. Comput. Vis. Pattern Recog.}, 2020.

\bibitem[Brachmann et~al.(2023)Brachmann, Cavallari, and
  Prisacariu]{brachmann2023ace}
Eric Brachmann, Tommaso Cavallari, and Victor~Adrian Prisacariu.
\newblock Accelerated coordinate encoding: Learning to relocalize in minutes
  using {RGB} and poses.
\newblock In \emph{IEEE Conf. Comput. Vis. Pattern Recog.}, 2023.

\bibitem[Brachmann et~al.(2024)Brachmann, Wynn, Chen, Cavallari, Monszpart,
  Turmukhambetov, and Prisacariu]{brachmann2024acezero}
Eric Brachmann, Jamie Wynn, Shuai Chen, Tommaso Cavallari, {\'{A}}ron
  Monszpart, Daniyar Turmukhambetov, and Victor~Adrian Prisacariu.
\newblock Scene coordinate reconstruction: Posing of image collections via
  incremental learning of a relocalizer.
\newblock In \emph{Eur. Conf. Comput. Vis.}, 2024.

\bibitem[Butler et~al.(2012)Butler, Wulff, Stanley, and
  Black]{butler2012naturalistic}
Daniel~J Butler, Jonas Wulff, Garrett~B Stanley, and Michael~J Black.
\newblock A naturalistic open source movie for optical flow evaluation.
\newblock In \emph{ECCV}, 2012.

\bibitem[Campbell et~al.(2008)Campbell, Vogiatzis, Hern{\'a}ndez, and
  Cipolla]{campbell2008using}
Neill~DF Campbell, George Vogiatzis, Carlos Hern{\'a}ndez, and Roberto Cipolla.
\newblock Using multiple hypotheses to improve depth-maps for multi-view
  stereo.
\newblock In \emph{Eur. Conf. Comput. Vis.}, 2008.

\bibitem[Campos et~al.(2021)Campos, Elvira, Rodr{\'\i}guez, Montiel, and
  Tard{\'o}s]{campos2021orb}
Carlos Campos, Richard Elvira, Juan J~G{\'o}mez Rodr{\'\i}guez, Jos{\'e}~MM
  Montiel, and Juan~D Tard{\'o}s.
\newblock {ORB-SLAM3}: An accurate open-source library for visual,
  visual-inertial and multi-map {SLAM}.
\newblock \emph{IEEE Transactions on Robotics}, 2021.

\bibitem[Chang and Chen(2018)]{chang2018pyramid}
Jia-Ren Chang and Yong-Sheng Chen.
\newblock Pyramid stereo matching network.
\newblock In \emph{IEEE Conf. Comput. Vis. Pattern Recog.}, 2018.

\bibitem[Chen(2017)]{chen2017rethinking}
Liang-Chieh Chen.
\newblock Rethinking atrous convolution for semantic image segmentation.
\newblock \emph{arXiv preprint arXiv:1706.05587}, 2017.

\bibitem[Curless and Levoy(1996)]{curless1996volumetric}
Brian Curless and Marc Levoy.
\newblock A volumetric method for building complex models from range images.
\newblock In \emph{SIGGRAPH}, 1996.

\bibitem[Davison et~al.(2007)Davison, Reid, Molton, and
  Stasse]{davison2007monoslam}
Andrew~J Davison, Ian~D Reid, Nicholas~D Molton, and Olivier Stasse.
\newblock {MonoSLAM}: Real-time single camera {SLAM}.
\newblock \emph{IEEE Trans. Pattern Anal. Mach. Intell.}, 2007.

\bibitem[Doersch et~al.(2022)Doersch, Gupta, Markeeva, Recasens, Smaira, Aytar,
  Carreira, Zisserman, and Yang]{doersch2022tap}
Carl Doersch, Ankush Gupta, Larisa Markeeva, Adria Recasens, Lucas Smaira,
  Yusuf Aytar, Joao Carreira, Andrew Zisserman, and Yi Yang.
\newblock {TAP-Vid}: A benchmark for tracking any point in a video.
\newblock \emph{NeurIPS}, 2022.

\bibitem[Doersch et~al.(2024)Doersch, Luc, Yang, Gokay, Koppula, Gupta,
  Heyward, Rocco, Goroshin, Carreira, and Zisserman]{doersch2024bootstap}
Carl Doersch, Pauline Luc, Yi Yang, Dilara Gokay, Skanda Koppula, Ankush Gupta,
  Joseph Heyward, Ignacio Rocco, Ross Goroshin, João Carreira, and Andrew
  Zisserman.
\newblock {BootsTAP}: Bootstrapped training for tracking any point.
\newblock \emph{ICCV}, 2024.

\bibitem[Dosovitskiy(2021)]{dosovitskiy2020image}
Alexey Dosovitskiy.
\newblock An image is worth 16x16 words: Transformers for image recognition at
  scale.
\newblock \emph{ICLR}, 2021.

\bibitem[Dosovitskiy et~al.(2015)Dosovitskiy, Fischer, Ilg, Hausser, Hazirbas,
  Golkov, Van Der~Smagt, Cremers, and Brox]{dosovitskiy2015flownet}
Alexey Dosovitskiy, Philipp Fischer, Eddy Ilg, Philip Hausser, Caner Hazirbas,
  Vladimir Golkov, Patrick Van Der~Smagt, Daniel Cremers, and Thomas Brox.
\newblock {FlowNet}: Learning optical flow with convolutional networks.
\newblock In \emph{Int. Conf. Comput. Vis.}, 2015.

\bibitem[Engel et~al.(2017)Engel, Koltun, and Cremers]{engel2017direct}
Jakob Engel, Vladlen Koltun, and Daniel Cremers.
\newblock Direct sparse odometry.
\newblock \emph{IEEE Trans. Pattern Anal. Mach. Intell.}, 2017.

\bibitem[Fu et~al.(2024)Fu, Liu, Kulkarni, Kautz, Efros, and
  Wang]{Fu_2024_CVPR}
Yang Fu, Sifei Liu, Amey Kulkarni, Jan Kautz, Alexei~A. Efros, and Xiaolong
  Wang.
\newblock {COLMAP-Free} {3D} gaussian splatting.
\newblock In \emph{IEEE Conf. Comput. Vis. Pattern Recog.}, 2024.

\bibitem[Furukawa and Ponce(2009)]{furukawa2009accurate}
Yasutaka Furukawa and Jean Ponce.
\newblock Accurate, dense, and robust multiview stereopsis.
\newblock \emph{IEEE Trans. Pattern Anal. Mach. Intell.}, 2009.

\bibitem[Furukawa et~al.(2010)Furukawa, Curless, Seitz, and
  Szeliski]{furukawa2010towards}
Yasutaka Furukawa, Brian Curless, Steven~M Seitz, and Richard Szeliski.
\newblock Towards internet-scale multi-view stereo.
\newblock In \emph{IEEE Conf. Comput. Vis. Pattern Recog.}, 2010.

\bibitem[Galliani et~al.(2015)Galliani, Lasinger, and
  Schindler]{galliani2015massively}
Silvano Galliani, Katrin Lasinger, and Konrad Schindler.
\newblock Massively parallel multiview stereopsis by surface normal diffusion.
\newblock In \emph{Int. Conf. Comput. Vis.}, 2015.

\bibitem[Gao et~al.(2022)Gao, Li, Tulsiani, Russell, and
  Kanazawa]{gao2022monocular}
Hang Gao, Ruilong Li, Shubham Tulsiani, Bryan Russell, and Angjoo Kanazawa.
\newblock Monocular dynamic view synthesis: A reality check.
\newblock \emph{NeurIPS}, 2022.

\bibitem[Gao et~al.(2024)Gao, Holynski, Henzler, Brussee, Martin-Brualla,
  Srinivasan, Barron, and Poole]{gao2024cat3d}
Ruiqi Gao, Aleksander Holynski, Philipp Henzler, Arthur Brussee, Ricardo
  Martin-Brualla, Pratul Srinivasan, Jonathan~T Barron, and Ben Poole.
\newblock {CAT3D}: Create anything in {3D} with multi-view diffusion models.
\newblock \emph{NeurIPS}, 2024.

\bibitem[Geiger et~al.(2013)Geiger, Lenz, Stiller, and
  Urtasun]{geiger2013vision}
Andreas Geiger, Philip Lenz, Christoph Stiller, and Raquel Urtasun.
\newblock Vision meets robotics: The {KITTI} dataset.
\newblock \emph{IJRR}, 2013.

\bibitem[Grauman et~al.(2024)Grauman, Westbury, Torresani, Kitani, Malik,
  Afouras, Ashutosh, Baiyya, Bansal, Boote, Byrne, Chavis, Chen, Cheng, Chu,
  Crane, Dasgupta, Dong, Escobar, Forigua, Gebreselasie, Haresh, Huang, Islam,
  Jain, Khirodkar, Kukreja, Liang, Liu, Majumder, Mao, Martin, Mavroudi,
  Nagarajan, Ragusa, Ramakrishnan, Seminara, Somayazulu, Song, Su, Xue, Zhang,
  Zhang, Castillo, Chen, Fu, Furuta, Gonzalez, Gupta, Hu, Huang, Huang, Khoo,
  Kumar, Kuo, Lakhavani, Liu, Luo, Luo, Meredith, Miller, Oguntola, Pan, Peng,
  Pramanick, Ramazanova, Ryan, Shan, Somasundaram, Song, Southerland, Tateno,
  Wang, Wang, Yagi, Yan, Yang, Yu, Zha, Zhao, Zhao, Zhu, Zhuo, Arbelaez,
  Bertasius, Damen, Engel, Farinella, Furnari, Ghanem, Hoffman, Jawahar,
  Newcombe, Park, Rehg, Sato, Savva, Shi, Shou, and Wray]{Grauman_2024_CVPR}
Kristen Grauman, Andrew Westbury, Lorenzo Torresani, Kris Kitani, Jitendra
  Malik, Triantafyllos Afouras, Kumar Ashutosh, Vijay Baiyya, Siddhant Bansal,
  Bikram Boote, Eugene Byrne, Zach Chavis, Joya Chen, Feng Cheng, Fu-Jen Chu,
  Sean Crane, Avijit Dasgupta, Jing Dong, Maria Escobar, Cristhian Forigua,
  Abrham Gebreselasie, Sanjay Haresh, Jing Huang, Md~Mohaiminul Islam, Suyog
  Jain, Rawal Khirodkar, Devansh Kukreja, Kevin~J Liang, Jia-Wei Liu, Sagnik
  Majumder, Yongsen Mao, Miguel Martin, Effrosyni Mavroudi, Tushar Nagarajan,
  Francesco Ragusa, Santhosh~Kumar Ramakrishnan, Luigi Seminara, Arjun
  Somayazulu, Yale Song, Shan Su, Zihui Xue, Edward Zhang, Jinxu Zhang, Angela
  Castillo, Changan Chen, Xinzhu Fu, Ryosuke Furuta, Cristina Gonzalez, Prince
  Gupta, Jiabo Hu, Yifei Huang, Yiming Huang, Weslie Khoo, Anush Kumar, Robert
  Kuo, Sach Lakhavani, Miao Liu, Mi Luo, Zhengyi Luo, Brighid Meredith, Austin
  Miller, Oluwatumininu Oguntola, Xiaqing Pan, Penny Peng, Shraman Pramanick,
  Merey Ramazanova, Fiona Ryan, Wei Shan, Kiran Somasundaram, Chenan Song,
  Audrey Southerland, Masatoshi Tateno, Huiyu Wang, Yuchen Wang, Takuma Yagi,
  Mingfei Yan, Xitong Yang, Zecheng Yu, Shengxin~Cindy Zha, Chen Zhao, Ziwei
  Zhao, Zhifan Zhu, Jeff Zhuo, Pablo Arbelaez, Gedas Bertasius, Dima Damen,
  Jakob Engel, Giovanni~Maria Farinella, Antonino Furnari, Bernard Ghanem, Judy
  Hoffman, C.V. Jawahar, Richard Newcombe, Hyun~Soo Park, James~M. Rehg, Yoichi
  Sato, Manolis Savva, Jianbo Shi, Mike~Zheng Shou, and Michael Wray.
\newblock {Ego-Exo4D}: Understanding skilled human activity from first- and
  third-person perspectives.
\newblock In \emph{IEEE Conf. Comput. Vis. Pattern Recog.}, 2024.

\bibitem[Greff et~al.(2022)Greff, Belletti, Beyer, Doersch, Du, Duckworth,
  Fleet, Gnanapragasam, Golemo, Herrmann, Kipf, Kundu, Lagun, Laradji, Liu,
  Meyer, Miao, Nowrouzezahrai, Oztireli, Pot, Radwan, Rebain, Sabour, Sajjadi,
  Sela, Sitzmann, Stone, Sun, Vora, Wang, Wu, Yi, Zhong, and
  Tagliasacchi]{greff2021kubric}
Klaus Greff, Francois Belletti, Lucas Beyer, Carl Doersch, Yilun Du, Daniel
  Duckworth, David~J Fleet, Dan Gnanapragasam, Florian Golemo, Charles
  Herrmann, Thomas Kipf, Abhijit Kundu, Dmitry Lagun, Issam Laradji,
  Hsueh-Ti~(Derek) Liu, Henning Meyer, Yishu Miao, Derek Nowrouzezahrai, Cengiz
  Oztireli, Etienne Pot, Noha Radwan, Daniel Rebain, Sara Sabour, Mehdi S.~M.
  Sajjadi, Matan Sela, Vincent Sitzmann, Austin Stone, Deqing Sun, Suhani Vora,
  Ziyu Wang, Tianhao Wu, Kwang~Moo Yi, Fangcheng Zhong, and Andrea
  Tagliasacchi.
\newblock Kubric: a scalable dataset generator.
\newblock In \emph{IEEE Conf. Comput. Vis. Pattern Recog.}, 2022.

\bibitem[Harley et~al.(2022)Harley, Fang, and Fragkiadaki]{harley2022particle}
Adam~W Harley, Zhaoyuan Fang, and Katerina Fragkiadaki.
\newblock Particle video revisited: Tracking through occlusions using point
  trajectories.
\newblock In \emph{Eur. Conf. Comput. Vis.}, 2022.

\bibitem[Hirschm{\"u}ller et~al.(2002)Hirschm{\"u}ller, Innocent, and
  Garibaldi]{hirschmuller2002real}
Heiko Hirschm{\"u}ller, Peter~R Innocent, and Jon Garibaldi.
\newblock Real-time correlation-based stereo vision with reduced border errors.
\newblock \emph{IJCV}, 2002.

\bibitem[Holynski et~al.(2020)Holynski, Geraghty, Frahm, Sweeney, and
  Szeliski]{holynski2020reducing}
Aleksander Holynski, David Geraghty, Jan-Michael Frahm, Chris Sweeney, and
  Richard Szeliski.
\newblock Reducing drift in structure from motion using extended features.
\newblock In \emph{3DV}, 2020.

\bibitem[Hoppe et~al.(1992)Hoppe, DeRose, Duchamp, McDonald, and
  Stuetzle]{hoppe1992surface}
Hugues Hoppe, Tony DeRose, Tom Duchamp, John McDonald, and Werner Stuetzle.
\newblock Surface reconstruction from unorganized points.
\newblock In \emph{SIGGRAPH}, 1992.

\bibitem[Hu et~al.(2024)Hu, Gao, Li, Zhao, Cun, Zhang, Quan, and
  Shan]{hu2024depthcrafter}
Wenbo Hu, Xiangjun Gao, Xiaoyu Li, Sijie Zhao, Xiaodong Cun, Yong Zhang, Long
  Quan, and Ying Shan.
\newblock {DepthCrafter}: Generating consistent long depth sequences for
  open-world videos.
\newblock \emph{arXiv preprint arXiv:2409.02095}, 2024.

\bibitem[I\c{s}{\i}k et~al.(2023)I\c{s}{\i}k, Rünz, Georgopoulos, Khakhulin,
  Starck, Agapito, and Nießner]{isik2023humanrf}
Mustafa I\c{s}{\i}k, Martin Rünz, Markos Georgopoulos, Taras Khakhulin,
  Jonathan Starck, Lourdes Agapito, and Matthias Nießner.
\newblock {HumanRF}: High-fidelity neural radiance fields for humans in motion.
\newblock \emph{ACM Transactions on Graphics (TOG)}, 2023.

\bibitem[Jancosek and Pajdla(2011)]{jancosek2011multi}
Michal Jancosek and Tomas Pajdla.
\newblock Multi-view reconstruction preserving weakly-supported surfaces.
\newblock In \emph{IEEE Conf. Comput. Vis. Pattern Recog.}, 2011.

\bibitem[Jing et~al.(2025)Jing, Mao, and
  Mikolajczyk]{jing2024matchstereovideos}
Junpeng Jing, Ye Mao, and Krystian Mikolajczyk.
\newblock Match-stereo-videos: Bidirectional alignment for consistent dynamic
  stereo matching.
\newblock In \emph{ECCV}, 2025.

\bibitem[Joo et~al.(2015)Joo, Liu, Tan, Gui, Nabbe, Matthews, Kanade, Nobuhara,
  and Sheikh]{Joo_2015_ICCV}
Hanbyul Joo, Hao Liu, Lei Tan, Lin Gui, Bart Nabbe, Iain Matthews, Takeo
  Kanade, Shohei Nobuhara, and Yaser Sheikh.
\newblock Panoptic studio: A massively multiview system for social motion
  capture.
\newblock In \emph{Int. Conf. Comput. Vis.}, 2015.

\bibitem[Karaev et~al.(2023)Karaev, Rocco, Graham, Neverova, Vedaldi, and
  Rupprecht]{karaev2023dynamicstereo}
Nikita Karaev, Ignacio Rocco, Benjamin Graham, Natalia Neverova, Andrea
  Vedaldi, and Christian Rupprecht.
\newblock {DynamicStereo}: Consistent dynamic depth from stereo videos.
\newblock In \emph{IEEE Conf. Comput. Vis. Pattern Recog.}, 2023.

\bibitem[Kazhdan et~al.(2006)Kazhdan, Bolitho, and Hoppe]{kazhdan2006poisson}
Michael Kazhdan, Matthew Bolitho, and Hugues Hoppe.
\newblock Poisson surface reconstruction.
\newblock In \emph{Proceedings of the fourth Eurographics symposium on Geometry
  processing}, 2006.

\bibitem[Ke et~al.(2024)Ke, Obukhov, Huang, Metzger, Daudt, and
  Schindler]{ke2024repurposing}
Bingxin Ke, Anton Obukhov, Shengyu Huang, Nando Metzger, Rodrigo~Caye Daudt,
  and Konrad Schindler.
\newblock Repurposing diffusion-based image generators for monocular depth
  estimation.
\newblock In \emph{IEEE Conf. Comput. Vis. Pattern Recog.}, 2024.

\bibitem[Kendall et~al.(2017)Kendall, Martirosyan, Dasgupta, Henry, Kennedy,
  Bachrach, and Bry]{kendall2017end}
Alex Kendall, Hayk Martirosyan, Saumitro Dasgupta, Peter Henry, Ryan Kennedy,
  Abraham Bachrach, and Adam Bry.
\newblock End-to-end learning of geometry and context for deep stereo
  regression.
\newblock In \emph{Int. Conf. Comput. Vis.}, 2017.

\bibitem[Kirschstein et~al.(2023)Kirschstein, Qian, Giebenhain, Walter, and
  Nie\ss{}ner]{kirschstein2023nersemble}
Tobias Kirschstein, Shenhan Qian, Simon Giebenhain, Tim Walter, and Matthias
  Nie\ss{}ner.
\newblock {NeRSemble}: Multi-view radiance field reconstruction of human heads.
\newblock \emph{ACM Trans. Graph.}, 2023.

\bibitem[Klaus et~al.(2006)Klaus, Sormann, and Karner]{klaus2006segment}
Andreas Klaus, Mario Sormann, and Konrad Karner.
\newblock Segment-based stereo matching using belief propagation and a
  self-adapting dissimilarity measure.
\newblock In \emph{ICPR}, 2006.

\bibitem[Kopf et~al.(2021)Kopf, Rong, and Huang]{kopf2021rcvd}
Johannes Kopf, Xuejian Rong, and Jia-Bin Huang.
\newblock Robust consistent video depth estimation.
\newblock In \emph{IEEE Conf. Comput. Vis. Pattern Recog.}, 2021.

\bibitem[Koppula et~al.(2024)Koppula, Rocco, Yang, Heyward, Carreira,
  Zisserman, Brostow, and Doersch]{koppula2024tapvid3d}
Skanda Koppula, Ignacio Rocco, Yi Yang, Joe Heyward, João Carreira, Andrew
  Zisserman, Gabriel Brostow, and Carl Doersch.
\newblock {TAPVid-3D}: A benchmark for tracking any point in {3D}.
\newblock In \emph{NeurIPS}, 2024.

\bibitem[Lei et~al.(2024)Lei, Weng, Harley, Guibas, and
  Daniilidis]{lei2024mosca}
Jiahui Lei, Yijia Weng, Adam Harley, Leonidas Guibas, and Kostas Daniilidis.
\newblock {MoSca}: Dynamic gaussian fusion from casual videos via {4D} motion
  scaffolds.
\newblock \emph{arXiv preprint arXiv:2405.17421}, 2024.

\bibitem[Leroy et~al.(2024)Leroy, Cabon, and Revaud]{leroy2024grounding}
Vincent Leroy, Yohann Cabon, and J{\'e}r{\^o}me Revaud.
\newblock Grounding image matching in {3D} with {MASt3R}.
\newblock In \emph{ECCV}, 2024.

\bibitem[Li and Snavely(2018)]{li2018megadepth}
Zhengqi Li and Noah Snavely.
\newblock {MegaDepth}: Learning single-view depth prediction from internet
  photos.
\newblock In \emph{IEEE Conf. Comput. Vis. Pattern Recog.}, 2018.

\bibitem[Li et~al.(2019)Li, Dekel, Cole, Tucker, Snavely, Liu, and
  Freeman]{li2019learning}
Zhengqi Li, Tali Dekel, Forrester Cole, Richard Tucker, Noah Snavely, Ce Liu,
  and William~T Freeman.
\newblock Learning the depths of moving people by watching frozen people.
\newblock In \emph{IEEE Conf. Comput. Vis. Pattern Recog.}, 2019.

\bibitem[Li et~al.(2021)Li, Niklaus, Snavely, and Wang]{li2021neural}
Zhengqi Li, Simon Niklaus, Noah Snavely, and Oliver Wang.
\newblock Neural scene flow fields for space-time view synthesis of dynamic
  scenes.
\newblock In \emph{IEEE Conf. Comput. Vis. Pattern Recog.}, 2021.

\bibitem[Li et~al.(2023{\natexlab{a}})Li, Wang, Cole, Tucker, and
  Snavely]{li2023dynibar}
Zhengqi Li, Qianqian Wang, Forrester Cole, Richard Tucker, and Noah Snavely.
\newblock {DynIBaR}: Neural dynamic image-based rendering.
\newblock In \emph{IEEE Conf. Comput. Vis. Pattern Recog.}, 2023{\natexlab{a}}.

\bibitem[Li et~al.(2023{\natexlab{b}})Li, Ye, Wang, Creighton, Taylor,
  Venkatesh, and Unberath]{li2023temporally}
Zhaoshuo Li, Wei Ye, Dilin Wang, Francis~X Creighton, Russell~H Taylor, Ganesh
  Venkatesh, and Mathias Unberath.
\newblock Temporally consistent online depth estimation in dynamic scenes.
\newblock In \emph{IEEE Conf. Comput. Vis. Pattern Recog.}, 2023{\natexlab{b}}.

\bibitem[Li et~al.(2024)Li, Tucker, Cole, Wang, Jin, Ye, Kanazawa, Holynski,
  and Snavely]{li2024megasam}
Zhengqi Li, Richard Tucker, Forrester Cole, Qianqian Wang, Linyi Jin, Vickie
  Ye, Angjoo Kanazawa, Aleksander Holynski, and Noah Snavely.
\newblock {MegaSaM}: Accurate, fast, and robust structure and motion from
  casual dynamic videos.
\newblock \emph{arXiv preprint arXiv:2412.04463}, 2024.

\bibitem[Lin et~al.(2021)Lin, Ma, Torralba, and Lucey]{lin2021barf}
Chen-Hsuan Lin, Wei-Chiu Ma, Antonio Torralba, and Simon Lucey.
\newblock {BARF}: Bundle-adjusting neural radiance fields.
\newblock In \emph{IEEE Conf. Comput. Vis. Pattern Recog.}, 2021.

\bibitem[Lindenberger et~al.(2021)Lindenberger, Sarlin, Larsson, and
  Pollefeys]{lindenberger2021pixsfm}
Philipp Lindenberger, Paul-Edouard Sarlin, Viktor Larsson, and Marc Pollefeys.
\newblock Pixel-perfect structure-from-motion with featuremetric refinement.
\newblock In \emph{ICCV}, 2021.

\bibitem[Liu et~al.(2023)Liu, Gao, Meuleman, Tseng, Saraf, Kim, Chuang, Kopf,
  and Huang]{liu2023robust}
Yu-Lun Liu, Chen Gao, Andreas Meuleman, Hung-Yu Tseng, Ayush Saraf, Changil
  Kim, Yung-Yu Chuang, Johannes Kopf, and Jia-Bin Huang.
\newblock Robust dynamic radiance fields.
\newblock In \emph{IEEE Conf. Comput. Vis. Pattern Recog.}, 2023.

\bibitem[Lowe(2004)]{lowe2004sift}
G Lowe.
\newblock {SIFT}-the scale invariant feature transform.
\newblock \emph{Int. J}, 2004.

\bibitem[Luo et~al.(2020)Luo, Huang, Szeliski, Matzen, and
  Kopf]{luo2020consistent}
Xuan Luo, Jia-Bin Huang, Richard Szeliski, Kevin Matzen, and Johannes Kopf.
\newblock Consistent video depth estimation.
\newblock \emph{ACM Transactions on Graphics (ToG)}, 2020.

\bibitem[Mayer et~al.(2016)Mayer, Ilg, Hausser, Fischer, Cremers, Dosovitskiy,
  and Brox]{mayer2016large}
Nikolaus Mayer, Eddy Ilg, Philip Hausser, Philipp Fischer, Daniel Cremers,
  Alexey Dosovitskiy, and Thomas Brox.
\newblock A large dataset to train convolutional networks for disparity,
  optical flow, and scene flow estimation.
\newblock In \emph{IEEE Conf. Comput. Vis. Pattern Recog.}, 2016.

\bibitem[Mur-Artal et~al.(2015)Mur-Artal, Montiel, and
  Tard\'os]{murartal2015orbslam}
Ra\'ul Mur-Artal, J.~M.~M. Montiel, and Juan~D. Tard\'os.
\newblock {ORB-SLAM}: a versatile and accurate monocular {SLAM} system.
\newblock \emph{IEEE Trans. on Robotics}, 2015.

\bibitem[Newcombe et~al.(2015)Newcombe, Fox, and
  Seitz]{newcombe2015dynamicfusion}
Richard~A Newcombe, Dieter Fox, and Steven~M Seitz.
\newblock {DynamicFusion}: Reconstruction and tracking of non-rigid scenes in
  real-time.
\newblock In \emph{IEEE Conf. Comput. Vis. Pattern Recog.}, 2015.

\bibitem[Palazzolo et~al.(2019)Palazzolo, Behley, Lottes, Gigu\`ere, and
  Stachniss]{palazzolo2019iros}
E. Palazzolo, J. Behley, P. Lottes, P. Gigu\`ere, and C. Stachniss.
\newblock {ReFusion: {3D} Reconstruction in Dynamic Environments for {RGB-D}
  Cameras Exploiting Residuals}.
\newblock In \emph{IROS}, 2019.

\bibitem[Pan et~al.(2023)Pan, Charron, Yang, Peters, Whelan, Kong, Parkhi,
  Newcombe, and Ren]{pan2023aria}
Xiaqing Pan, Nicholas Charron, Yongqian Yang, Scott Peters, Thomas Whelan, Chen
  Kong, Omkar Parkhi, Richard Newcombe, and Yuheng~Carl Ren.
\newblock Aria digital twin: A new benchmark dataset for egocentric 3d machine
  perception.
\newblock In \emph{ICCV}, 2023.

\bibitem[Pang et~al.(2017)Pang, Sun, Ren, Yang, and Yan]{pang2017cascade}
Jiahao Pang, Wenxiu Sun, Jimmy~SJ Ren, Chengxi Yang, and Qiong Yan.
\newblock Cascade residual learning: A two-stage convolutional neural network
  for stereo matching.
\newblock In \emph{Proc. CVPR Workshops}, 2017.

\bibitem[Park et~al.(2010)Park, Shiratori, Matthews, and Sheikh]{park20103d}
Hyun~Soo Park, Takaaki Shiratori, Iain Matthews, and Yaser Sheikh.
\newblock {3D} reconstruction of a moving point from a series of {2D}
  projections.
\newblock In \emph{Eur. Conf. Comput. Vis.}, 2010.

\bibitem[Park et~al.(2021{\natexlab{a}})Park, Sinha, Barron, Bouaziz, Goldman,
  Seitz, and Martin-Brualla]{park2021nerfies}
Keunhong Park, Utkarsh Sinha, Jonathan~T Barron, Sofien Bouaziz, Dan~B Goldman,
  Steven~M Seitz, and Ricardo Martin-Brualla.
\newblock Nerfies: Deformable neural radiance fields.
\newblock In \emph{IEEE Conf. Comput. Vis. Pattern Recog.}, 2021{\natexlab{a}}.

\bibitem[Park et~al.(2021{\natexlab{b}})Park, Sinha, Hedman, Barron, Bouaziz,
  Goldman, Martin-Brualla, and Seitz]{park2021hypernerf}
Keunhong Park, Utkarsh Sinha, Peter Hedman, Jonathan~T Barron, Sofien Bouaziz,
  Dan~B Goldman, Ricardo Martin-Brualla, and Steven~M Seitz.
\newblock {HyperNeRF}: A higher-dimensional representation for topologically
  varying neural radiance fields.
\newblock \emph{arXiv preprint arXiv:2106.13228}, 2021{\natexlab{b}}.

\bibitem[Park et~al.(2023)Park, Henzler, Mildenhall, Barron, and
  Martin-Brualla]{park2023camp}
Keunhong Park, Philipp Henzler, Ben Mildenhall, Jonathan~T Barron, and Ricardo
  Martin-Brualla.
\newblock {CamP}: Camera preconditioning for neural radiance fields.
\newblock \emph{ACM Transactions on Graphics (TOG)}, 2023.

\bibitem[Piccinelli et~al.(2024)Piccinelli, Yang, Sakaridis, Segu, Li,
  Van~Gool, and Yu]{piccinelli2024unidepth}
Luigi Piccinelli, Yung-Hsu Yang, Christos Sakaridis, Mattia Segu, Siyuan Li,
  Luc Van~Gool, and Fisher Yu.
\newblock {U}ni{D}epth: Universal monocular metric depth estimation.
\newblock In \emph{IEEE Conf. Comput. Vis. Pattern Recog.}, 2024.

\bibitem[Pollefeys et~al.(2004)Pollefeys, Van~Gool, Vergauwen, Verbiest,
  Cornelis, Tops, and Koch]{pollefeys2004visual}
Marc Pollefeys, Luc Van~Gool, Maarten Vergauwen, Frank Verbiest, Kurt Cornelis,
  Jan Tops, and Reinhard Koch.
\newblock Visual modeling with a hand-held camera.
\newblock \emph{IJCV}, 2004.

\bibitem[Pollefeys et~al.(2008)Pollefeys, Nist{\'e}r, Frahm, Akbarzadeh,
  Mordohai, Clipp, Engels, Gallup, Kim, Merrell, et~al.]{pollefeys2008detailed}
Marc Pollefeys, David Nist{\'e}r, J-M Frahm, Amir Akbarzadeh, Philippos
  Mordohai, Brian Clipp, Chris Engels, David Gallup, S-J Kim, Paul Merrell,
  et~al.
\newblock Detailed real-time urban {3D} reconstruction from video.
\newblock \emph{IJCV}, 2008.

\bibitem[Polyak et~al.(2024)Polyak, Zohar, Brown, Tjandra, Sinha, Lee, Vyas,
  Shi, Ma, Chuang, et~al.]{polyak2024movie}
Adam Polyak, Amit Zohar, Andrew Brown, Andros Tjandra, Animesh Sinha, Ann Lee,
  Apoorv Vyas, Bowen Shi, Chih-Yao Ma, Ching-Yao Chuang, et~al.
\newblock {Movie Gen}: A cast of media foundation models.
\newblock \emph{arXiv preprint arXiv:2410.13720}, 2024.

\bibitem[Ranftl et~al.(2020)Ranftl, Lasinger, Hafner, Schindler, and
  Koltun]{ranftl2020towards}
Ren{\'e} Ranftl, Katrin Lasinger, David Hafner, Konrad Schindler, and Vladlen
  Koltun.
\newblock Towards robust monocular depth estimation: Mixing datasets for
  zero-shot cross-dataset transfer.
\newblock \emph{IEEE Trans. Pattern Anal. Mach. Intell.}, 2020.

\bibitem[Ranftl et~al.(2021)Ranftl, Bochkovskiy, and Koltun]{ranftl2021vision}
Ren{\'e} Ranftl, Alexey Bochkovskiy, and Vladlen Koltun.
\newblock Vision transformers for dense prediction.
\newblock In \emph{Int. Conf. Comput. Vis.}, 2021.

\bibitem[Schonberger and Frahm(2016)]{schonberger2016structure}
Johannes~L Schonberger and Jan-Michael Frahm.
\newblock Structure-from-motion revisited.
\newblock In \emph{IEEE Conf. Comput. Vis. Pattern Recog.}, 2016.

\bibitem[Sch{\"o}nberger et~al.(2016)Sch{\"o}nberger, Zheng, Frahm, and
  Pollefeys]{schonberger2016pixelwise}
Johannes~L Sch{\"o}nberger, Enliang Zheng, Jan-Michael Frahm, and Marc
  Pollefeys.
\newblock Pixelwise view selection for unstructured multi-view stereo.
\newblock In \emph{ECCV}, 2016.

\bibitem[Shao et~al.(2024)Shao, Yang, Zhou, Zhang, Shen, Poggi, and
  Liao]{shao2024learning}
Jiahao Shao, Yuanbo Yang, Hongyu Zhou, Youmin Zhang, Yujun Shen, Matteo Poggi,
  and Yiyi Liao.
\newblock Learning temporally consistent video depth from video diffusion
  priors.
\newblock \emph{arXiv preprint arXiv:2406.01493}, 2024.

\bibitem[Shen et~al.(2023)Shen, Cai, Wang, and Scherer]{shen2023dytanvo}
Shihao Shen, Yilin Cai, Wenshan Wang, and Sebastian Scherer.
\newblock {DytanVO}: Joint refinement of visual odometry and motion
  segmentation in dynamic environments.
\newblock In \emph{ICRA}, 2023.

\bibitem[Shih et~al.(2024)Shih, Ma, Boyice, Holynski, Cole, Curless, and
  Kontkanen]{shih2024extranerf}
Meng-Li Shih, Wei-Chiu Ma, Lorenzo Boyice, Aleksander Holynski, Forrester Cole,
  Brian Curless, and Janne Kontkanen.
\newblock {ExtraNeRF}: Visibility-aware view extrapolation of neural radiance
  fields with diffusion models.
\newblock In \emph{CVPR}, 2024.

\bibitem[Simon et~al.(2016)Simon, Valmadre, Matthews, and
  Sheikh]{simon2016kronecker}
Tomas Simon, Jack Valmadre, Iain Matthews, and Yaser Sheikh.
\newblock {Kronecker-Markov} prior for dynamic {3D} reconstruction.
\newblock \emph{IEEE Trans. Pattern Anal. Mach. Intell.}, 2016.

\bibitem[Snavely et~al.(2006)Snavely, Seitz, and Szeliski]{snavely2006photo}
Noah Snavely, Steven~M Seitz, and Richard Szeliski.
\newblock Photo tourism: exploring photo collections in {3D}.
\newblock In \emph{SIGGRAPH}, 2006.

\bibitem[Sun et~al.(2021)Sun, Vlasic, Herrmann, Jampani, Krainin, Chang, Zabih,
  Freeman, and Liu]{sun2021autoflow}
Deqing Sun, Daniel Vlasic, Charles Herrmann, Varun Jampani, Michael Krainin,
  Huiwen Chang, Ramin Zabih, William~T Freeman, and Ce Liu.
\newblock Autoflow: Learning a better training set for optical flow.
\newblock In \emph{IEEE Conf. Comput. Vis. Pattern Recog.}, 2021.

\bibitem[Sun et~al.(2022)Sun, Herrmann, Reda, Rubinstein, Fleet, and
  Freeman]{sun2022disentangling}
Deqing Sun, Charles Herrmann, Fitsum Reda, Michael Rubinstein, David~J Fleet,
  and William~T Freeman.
\newblock Disentangling architecture and training for optical flow.
\newblock In \emph{ECCV}, 2022.

\bibitem[Sun et~al.(2003)Sun, Zheng, and Shum]{sun2003stereo}
Jian Sun, Nan-Ning Zheng, and Heung-Yeung Shum.
\newblock Stereo matching using belief propagation.
\newblock \emph{IEEE Trans. Pattern Anal. Mach. Intell.}, 2003.

\bibitem[Sun et~al.(2020)Sun, Kretzschmar, Dotiwalla, Chouard, Patnaik, Tsui,
  Guo, Zhou, Chai, Caine, et~al.]{sun2020scalability}
Pei Sun, Henrik Kretzschmar, Xerxes Dotiwalla, Aurelien Chouard, Vijaysai
  Patnaik, Paul Tsui, James Guo, Yin Zhou, Yuning Chai, Benjamin Caine, et~al.
\newblock Scalability in perception for autonomous driving: {Waymo} open
  dataset.
\newblock In \emph{IEEE Conf. Comput. Vis. Pattern Recog.}, 2020.

\bibitem[Sweeney et~al.(2019)Sweeney, Holynski, Curless, and
  Seitz]{sweeney2019structure}
Chris Sweeney, Aleksander Holynski, Brian Curless, and Steve~M Seitz.
\newblock Structure from motion for panorama-style videos.
\newblock \emph{arXiv preprint arXiv:1906.03539}, 2019.

\bibitem[Tang and Tan(2018)]{tang2018ba}
Chengzhou Tang and Ping Tan.
\newblock {BA-Net}: Dense bundle adjustment network.
\newblock \emph{arXiv preprint arXiv:1806.04807}, 2018.

\bibitem[Team et~al.(2023)Team, Anil, Borgeaud, Alayrac, Yu, Soricut,
  Schalkwyk, Dai, Hauth, Millican, et~al.]{team2023gemini}
Gemini Team, Rohan Anil, Sebastian Borgeaud, Jean-Baptiste Alayrac, Jiahui Yu,
  Radu Soricut, Johan Schalkwyk, Andrew~M Dai, Anja Hauth, Katie Millican,
  et~al.
\newblock Gemini: a family of highly capable multimodal models.
\newblock \emph{arXiv preprint arXiv:2312.11805}, 2023.

\bibitem[Teed and Deng(2020)]{teed2020raft}
Zachary Teed and Jia Deng.
\newblock {RAFT}: Recurrent all-pairs field transforms for optical flow.
\newblock In \emph{ECCV}, 2020.

\bibitem[Teed and Deng(2021{\natexlab{a}})]{teed2021droid}
Zachary Teed and Jia Deng.
\newblock {DROID-SLAM: Deep Visual SLAM for Monocular, Stereo, and RGB-D
  Cameras}.
\newblock \emph{NeurIPS}, 2021{\natexlab{a}}.

\bibitem[Teed and Deng(2021{\natexlab{b}})]{teed2021raft3d}
Zachary Teed and Jia Deng.
\newblock {RAFT-3D}: Scene flow using rigid-motion embeddings.
\newblock In \emph{IEEE Conf. Comput. Vis. Pattern Recog.}, 2021{\natexlab{b}}.

\bibitem[Teed et~al.(2024)Teed, Lipson, and Deng]{teed2024deep}
Zachary Teed, Lahav Lipson, and Jia Deng.
\newblock Deep patch visual odometry.
\newblock \emph{NeurIPS}, 2024.

\bibitem[Van~Meerbergen et~al.(2002)Van~Meerbergen, Vergauwen, Pollefeys, and
  Van~Gool]{van2002hierarchical}
Geert Van~Meerbergen, Maarten Vergauwen, Marc Pollefeys, and Luc Van~Gool.
\newblock A hierarchical symmetric stereo algorithm using dynamic programming.
\newblock \emph{Int. J. Comput. Vis.}, 2002.

\bibitem[Vaswani(2017)]{vaswani2017attention}
A Vaswani.
\newblock Attention is all you need.
\newblock \emph{NeurIPS}, 2017.

\bibitem[Vo et~al.(2016)Vo, Narasimhan, and Sheikh]{vo2016spatiotemporal}
Minh Vo, Srinivasa~G Narasimhan, and Yaser Sheikh.
\newblock Spatiotemporal bundle adjustment for dynamic {3D} reconstruction.
\newblock In \emph{IEEE Conf. Comput. Vis. Pattern Recog.}, 2016.

\bibitem[Wang et~al.(2023{\natexlab{a}})Wang, Lucey, Perazzi, and
  Wang]{wang2019web}
Chaoyang Wang, Simon Lucey, Federico Perazzi, and Oliver Wang.
\newblock Web stereo video supervision for depth prediction from dynamic
  scenes, 2023{\natexlab{a}}.

\bibitem[Wang et~al.(2024{\natexlab{a}})Wang, Karaev, Rupprecht, and
  Novotny]{wang2023vggsfm}
Jianyuan Wang, Nikita Karaev, Christian Rupprecht, and David Novotny.
\newblock {VGGSfM}: Visual geometry grounded deep structure from motion.
\newblock In \emph{CVPR}, 2024{\natexlab{a}}.

\bibitem[Wang et~al.(2024{\natexlab{b}})Wang, Ye, Gao, Austin, Li, and
  Kanazawa]{wang2024shape}
Qianqian Wang, Vickie Ye, Hang Gao, Jake Austin, Zhengqi Li, and Angjoo
  Kanazawa.
\newblock Shape of motion: {4D} reconstruction from a single video.
\newblock \emph{arXiv preprint arXiv:2407.13764}, 2024{\natexlab{b}}.

\bibitem[Wang et~al.(2024{\natexlab{c}})Wang, Leroy, Cabon, Chidlovskii, and
  Revaud]{wang2024dust3r}
Shuzhe Wang, Vincent Leroy, Yohann Cabon, Boris Chidlovskii, and Jerome Revaud.
\newblock {DUSt3R}: Geometric {3D} vision made easy.
\newblock In \emph{IEEE Conf. Comput. Vis. Pattern Recog.}, 2024{\natexlab{c}}.

\bibitem[Wang et~al.(2023{\natexlab{b}})Wang, Shi, Li, Huang, Cao, Zhang, Xian,
  and Lin]{NVDS}
Yiran Wang, Min Shi, Jiaqi Li, Zihao Huang, Zhiguo Cao, Jianming Zhang, Ke
  Xian, and Guosheng Lin.
\newblock Neural video depth stabilizer.
\newblock In \emph{Int. Conf. Comput. Vis.}, 2023{\natexlab{b}}.

\bibitem[Wang et~al.(2025)Wang, Lipson, and Deng]{wang2025sea}
Yihan Wang, Lahav Lipson, and Jia Deng.
\newblock {SEA-RAFT}: Simple, efficient, accurate {RAFT} for optical flow.
\newblock In \emph{ECCV}, 2025.

\bibitem[Weber et~al.(2024)Weber, Holynski, Jampani, Saxena, Snavely, Kar, and
  Kanazawa]{weber2024nerfiller}
Ethan Weber, Aleksander Holynski, Varun Jampani, Saurabh Saxena, Noah Snavely,
  Abhishek Kar, and Angjoo Kanazawa.
\newblock Nerfiller: Completing scenes via generative {3D} inpainting.
\newblock In \emph{CVPR}, 2024.

\bibitem[Wu et~al.(2024)Wu, Gao, Poole, Trevithick, Zheng, Barron, and
  Holynski]{wu2024cat4d}
Rundi Wu, Ruiqi Gao, Ben Poole, Alex Trevithick, Changxi Zheng, Jonathan~T
  Barron, and Aleksander Holynski.
\newblock {CAT4D}: Create anything in {4D} with multi-view video diffusion
  models.
\newblock \emph{arXiv preprint arXiv:2411.18613}, 2024.

\bibitem[Yang et~al.(2024{\natexlab{a}})Yang, Kang, Huang, Xu, Feng, and
  Zhao]{depthanything}
Lihe Yang, Bingyi Kang, Zilong Huang, Xiaogang Xu, Jiashi Feng, and Hengshuang
  Zhao.
\newblock {Depth Anything}: Unleashing the power of large-scale unlabeled data.
\newblock In \emph{IEEE Conf. Comput. Vis. Pattern Recog.}, 2024{\natexlab{a}}.

\bibitem[Yang et~al.(2024{\natexlab{b}})Yang, Kang, Huang, Zhao, Xu, Feng, and
  Zhao]{yang2024depth}
Lihe Yang, Bingyi Kang, Zilong Huang, Zhen Zhao, Xiaogang Xu, Jiashi Feng, and
  Hengshuang Zhao.
\newblock {Depth Anything V2}.
\newblock \emph{NeurIPS}, 2024{\natexlab{b}}.

\bibitem[Yao et~al.(2018)Yao, Luo, Li, Fang, and Quan]{yao2018mvsnet}
Yao Yao, Zixin Luo, Shiwei Li, Tian Fang, and Long Quan.
\newblock {MVSNet}: Depth inference for unstructured multi-view stereo.
\newblock In \emph{Eur. Conf. Comput. Vis.}, 2018.

\bibitem[Yao et~al.(2019)Yao, Luo, Li, Shen, Fang, and Quan]{yao2019recurrent}
Yao Yao, Zixin Luo, Shiwei Li, Tianwei Shen, Tian Fang, and Long Quan.
\newblock Recurrent {MVSNet} for high-resolution multi-view stereo depth
  inference.
\newblock In \emph{IEEE Conf. Comput. Vis. Pattern Recog.}, 2019.

\bibitem[Yin et~al.(2021)Yin, Zhang, Wang, Niklaus, Mai, Chen, and
  Shen]{yin2021learning}
Wei Yin, Jianming Zhang, Oliver Wang, Simon Niklaus, Long Mai, Simon Chen, and
  Chunhua Shen.
\newblock Learning to recover {3D} scene shape from a single image.
\newblock In \emph{IEEE Conf. Comput. Vis. Pattern Recog.}, 2021.

\bibitem[Yin et~al.(2023)Yin, Zhang, Chen, Cai, Yu, Wang, Chen, and
  Shen]{yin2023metric3d}
Wei Yin, Chi Zhang, Hao Chen, Zhipeng Cai, Gang Yu, Kaixuan Wang, Xiaozhi Chen,
  and Chunhua Shen.
\newblock {Metric3D}: Towards zero-shot metric {3D} prediction from a single
  image.
\newblock In \emph{IEEE Conf. Comput. Vis. Pattern Recog.}, 2023.

\bibitem[Zhang et~al.(2019)Zhang, Prisacariu, Yang, and Torr]{zhang2019ga}
Feihu Zhang, Victor Prisacariu, Ruigang Yang, and Philip~HS Torr.
\newblock {GA-Net}: Guided aggregation net for end-to-end stereo matching.
\newblock In \emph{IEEE Conf. Comput. Vis. Pattern Recog.}, 2019.

\bibitem[Zhang et~al.(2025)Zhang, Herrmann, Hur, Jampani, Darrell, Cole, Sun,
  and Yang]{zhang2024monst3r}
Junyi Zhang, Charles Herrmann, Junhwa Hur, Varun Jampani, Trevor Darrell,
  Forrester Cole, Deqing Sun, and Ming-Hsuan Yang.
\newblock {MonST3R}: A simple approach for estimating geometry in the presence
  of motion.
\newblock \emph{ICLR}, 2025.

\bibitem[Zhang et~al.(2023)Zhang, Poggi, and
  Mattoccia]{zhang2023temporalstereo}
Youmin Zhang, Matteo Poggi, and Stefano Mattoccia.
\newblock {TemporalStereo}: Efficient spatial-temporal stereo matching network.
\newblock In \emph{IROS}, 2023.

\bibitem[Zhang et~al.(2021)Zhang, Cole, Tucker, Freeman, and
  Dekel]{zhang2021consistent}
Zhoutong Zhang, Forrester Cole, Richard Tucker, William~T Freeman, and Tali
  Dekel.
\newblock Consistent depth of moving objects in video.
\newblock \emph{ACM Transactions on Graphics (ToG)}, 2021.

\bibitem[Zhang et~al.(2022)Zhang, Cole, Li, Rubinstein, Snavely, and
  Freeman]{zhang2022structure}
Zhoutong Zhang, Forrester Cole, Zhengqi Li, Michael Rubinstein, Noah Snavely,
  and William~T Freeman.
\newblock Structure and motion from casual videos.
\newblock In \emph{Eur. Conf. Comput. Vis.}, 2022.

\bibitem[Zheng et~al.(2023)Zheng, Harley, Shen, Wetzstein, and
  Guibas]{zheng2023point}
Yang Zheng, Adam~W. Harley, Bokui Shen, Gordon Wetzstein, and Leonidas~J.
  Guibas.
\newblock {PointOdyssey}: A large-scale synthetic dataset for long-term point
  tracking.
\newblock In \emph{Int. Conf. Comput. Vis.}, 2023.

\bibitem[Zhou et~al.(2017)Zhou, Zhao, Puig, Fidler, Barriuso, and
  Torralba]{zhou2017scene}
Bolei Zhou, Hang Zhao, Xavier Puig, Sanja Fidler, Adela Barriuso, and Antonio
  Torralba.
\newblock Scene parsing through {ADE20K} dataset.
\newblock In \emph{CVPR}, 2017.

\bibitem[Zhou et~al.(2019)Zhou, Zhao, Puig, Xiao, Fidler, Barriuso, and
  Torralba]{zhou2019semantic}
Bolei Zhou, Hang Zhao, Xavier Puig, Tete Xiao, Sanja Fidler, Adela Barriuso,
  and Antonio Torralba.
\newblock Semantic understanding of scenes through the {ADE20K} dataset.
\newblock \emph{IJCV}, 2019.

\bibitem[Zhou et~al.(2018)Zhou, Tucker, Flynn, Fyffe, and
  Snavely]{zhou2018stereo}
Tinghui Zhou, Richard Tucker, John Flynn, Graham Fyffe, and Noah Snavely.
\newblock Stereo magnification: Learning view synthesis using multiplane
  images.
\newblock In \emph{SIGGRAPH}, 2018.

\end{thebibliography}
}

\clearpage

\setcounter{page}{1}
\maketitlesupplementary

\section{\dataset Statistics}
We collected around 110k clips from 6,493 Internet VR180 videos.
The videos were curated from public YouTube content using the tag ``VR180”, available under YouTube’s Standard License. We have released derived geometry, motion data, and video links, under a CC license \href{https://console.cloud.google.com/storage/browser/stereo4d;tab=objects?inv=1&invt=AbsjhA&prefix=&forceOnObjectsSortingFiltering=false}{here}.

\begin{figure}[t]
    \centering
    \includegraphics[width=\linewidth]{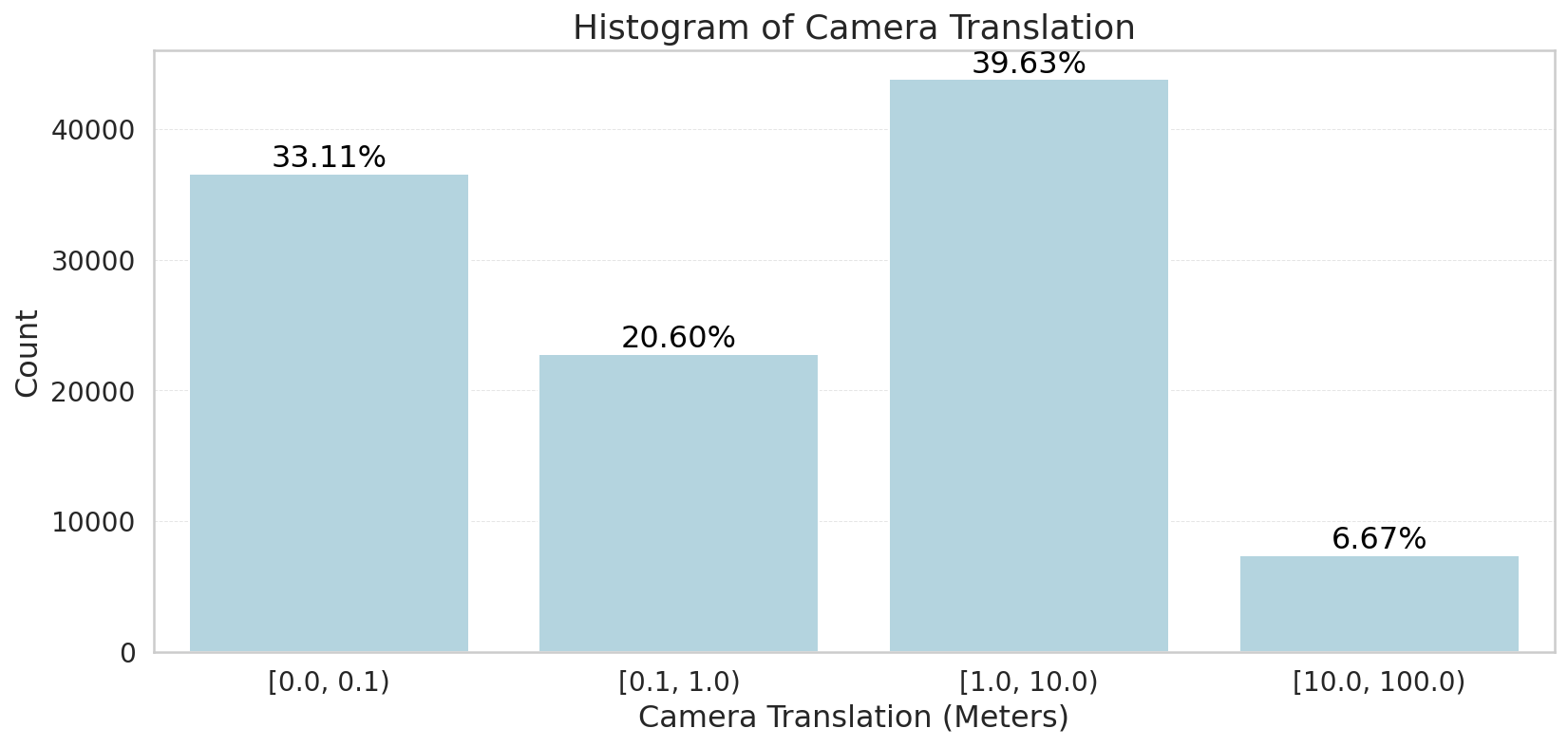}
    \caption{Camera statistics from \dataset. We measure the difference (in meters) of camera poses between the start and end frame of each video clip as calculated by SfM.}
    \label{fig:supp:camera_stats}
    \centering
    \includegraphics[width=\linewidth]{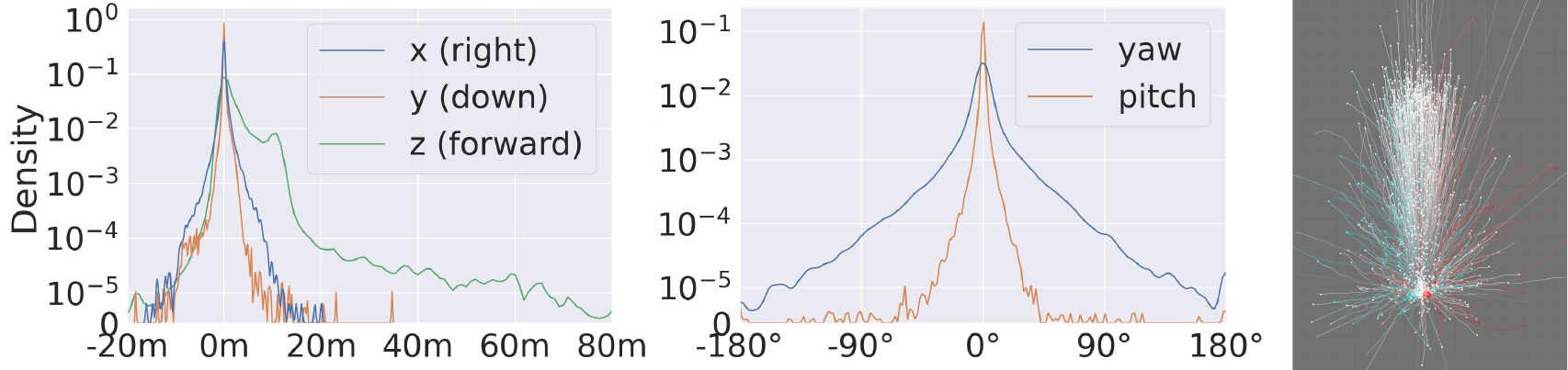}
    \caption{Camera statistics from \dataset. From left to right: a) the distribution of x,y,z camera translations (log scale), (b) the distribution of rotations, (c) a
sample of 5000 camera trajectories, viewed from above,
colored by final camera orientation (red=right, cyan=left)}
    \label{fig:supp:camera_stats_2}

    \centering
    \includegraphics[width=\linewidth]{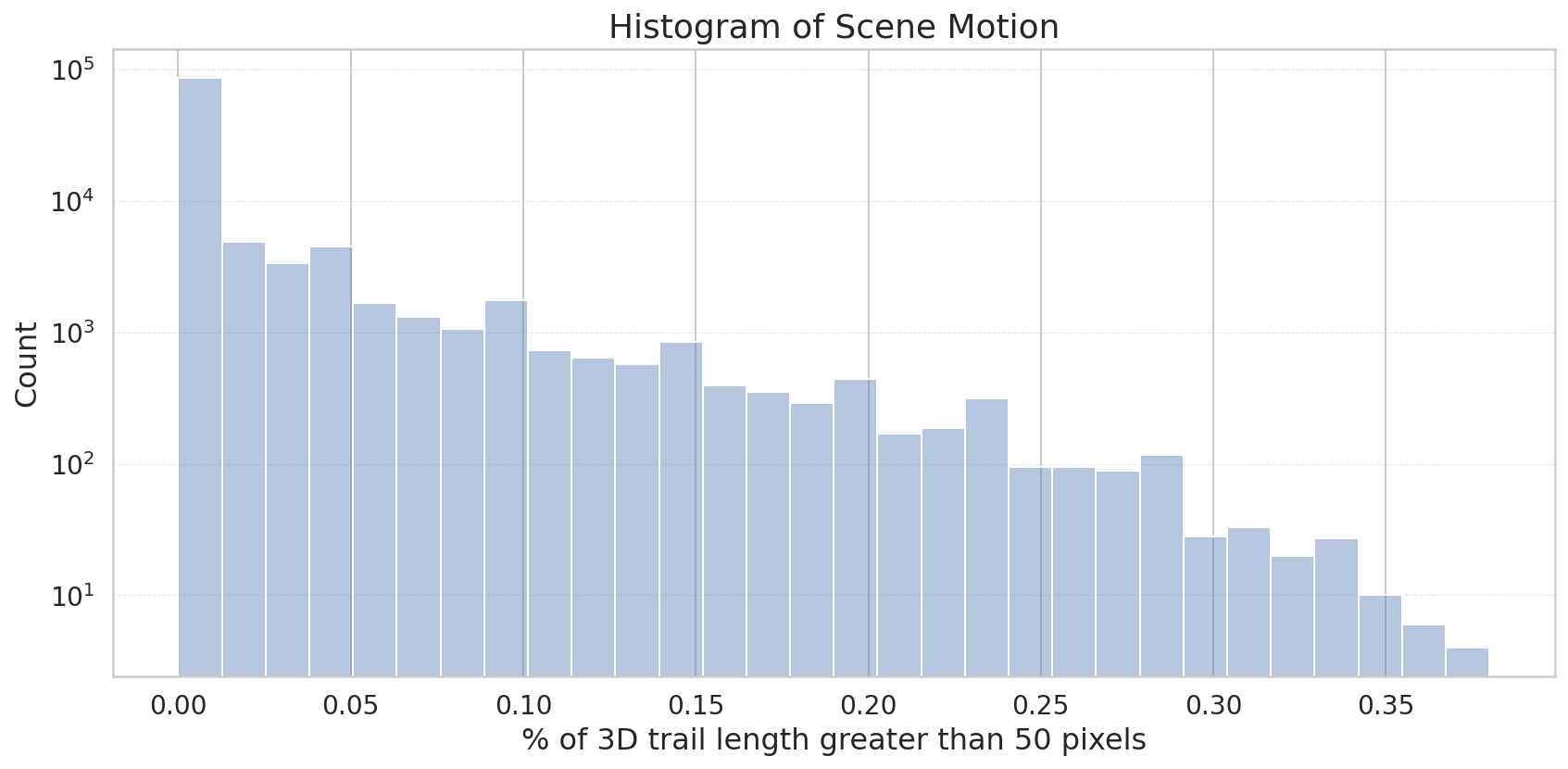}
    \caption{Scene motion statistics from \dataset. We measure scene motion in terms of pixel displacement projected onto the image frame. For each video, we measure the percentage of tracks that have 3D trail length greater than 50 pixels. The 3D trail length is measured by Eqn.~\ref{eqn:trail_length_def}. }
    \label{fig:supp:motion_stats}
\end{figure}

\Fig{supp:camera_stats} shows the camera translation distribution between the first and last frame of each clip. 
\Fig{supp:camera_stats_2} shows, from left to right: 
(a) the distribution of $x$,$y$,$z$ camera translations (log scale), 
(b) the distribution of rotations, 
(c) a sample of 5000 camera trajectories, viewed from above, colored by final camera orientation (red=right, cyan=left). 

In \Fig{supp:motion_stats}, we measure the motion in terms of pixel displacement projected onto the image frame. Measuring motion in pixel-space emphasizes motion that occurs closer to the camera, since such motion yields larger pixel displacements, while naturally de-emphasizing motion further from the camera. 

\section{Ablations}

\begin{figure}[t]
\centering\includegraphics[width=\linewidth]{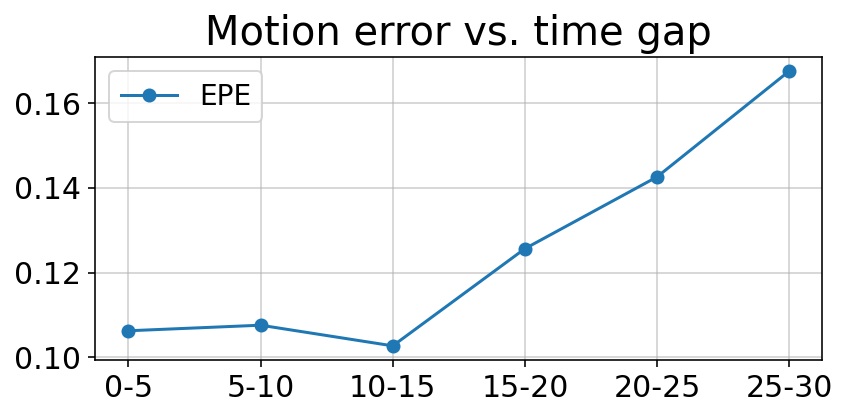}
    \caption{\textbf{Model performance with different input time gap}: Motion error vs. input frame gap. As the time gap increases, the motion magnitude and uncertainty grows, leading to an increase in error.}
    \label{fig:supp:epe_vs_time}
\end{figure}
\begin{figure*}[t]
    \centering
    \includegraphics[width=\linewidth]{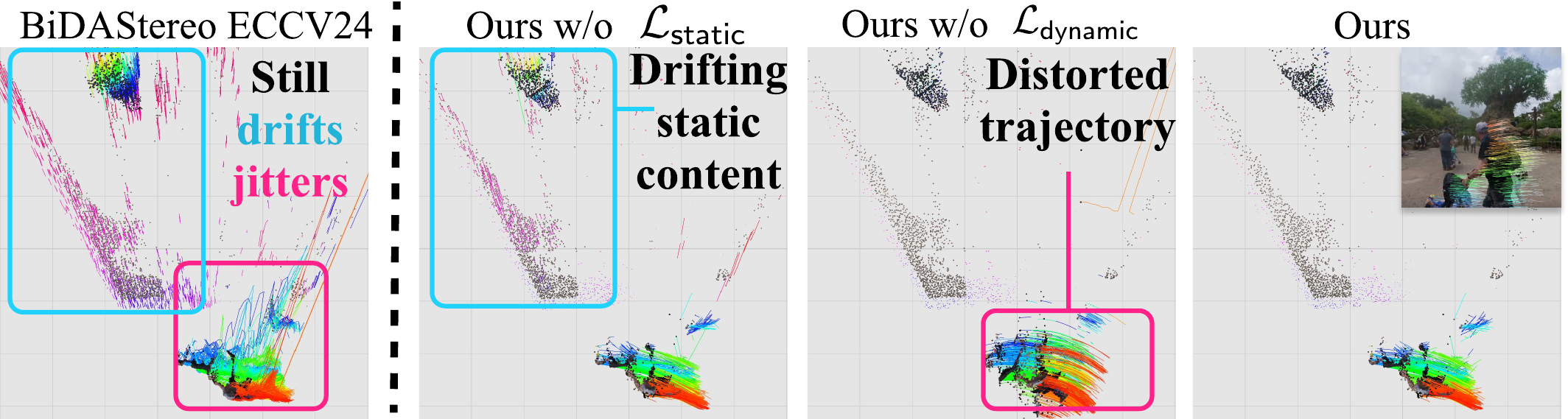}
    \caption{\textbf{Track optimization ablation}, extending Fig. 3. SOTA video stereo method, BiDAStereo~\cite{jing2024matchstereovideos}, still shows jitter and drift (left). 
Ablations (right): w/o $\mathcal{L}_{\mathsf{static}}$ static content reduces jitter from  $\mathcal{L}_{\mathsf{dynamic}}$ but still drifts; w/o $\mathcal{L}_{\mathsf{dynamic}}$ 
 dynamic trajectories are distorted along camera rays by $\mathcal{L}_{\mathsf{static}}$. }
    \label{fig:supp:track_ablation}
\end{figure*}

\noindent \bfpar{Track optimization.}
\Fig{supp:track_ablation} shows a top-down view of Fig 3. 
We tried multiple stereo depth methods when developing our system, e.g., BiDAStereo~\cite{jing2024matchstereovideos} but it still shows jitter and drift (left). 
That said, as more advanced stereo methods become available, we can adopt them. 
Ablations (right): w/o $\mathcal{L}_{\mathsf{static}}$ (static content reduces jitter from  $\mathcal{L}_{\mathsf{dynamic}}$ but still drifts); w/o $\mathcal{L}_{\mathsf{dynamic}}$ (dynamic trajectories are distorted along camera rays by $\mathcal{L}_{\mathsf{static}}$). 

\noindent \bfpar{Effect of time gap for \method.} 
We evaluate \method's capability with regard to time gap $\Delta_t = t_1 - t_0$ of input images. 
\Fig{supp:epe_vs_time} shows motion error vs.\ input frame gap, 
across 1k \dataset pairs. As the time gap increases, the motion magnitude and uncertainty grows, leading to an increase in error.

\noindent \bfpar{\method motion head ablation.} 
\method uses a separate motion head to predict motion for the pointmaps $M^{v\rightarrow t_q}$. Alternatively, one can also predict the deformed points with the same point head by conditioning on the time embedding. 
We compare 1) ours: predicting $M^{v\rightarrow t_q}$ with motion heads; and 2) directly regressing $P^{v\rightarrow t_q}$ with point heads and evaluate on the same Stereo4d test set in Table 1. Using a single head to predict motion (2) results in a drop in motion accuracy across all metrics: $\text{EPE}_\text{3D}$$\downarrow$ = $(\textbf{0.1110} \rightarrow 0.1401)$, $\delta_\text{3D}^{0.05}$$\uparrow$  = $(\textbf{65.07} \rightarrow 59.19)$, $\delta_\text{3D}^{0.10}$$\uparrow$  = $(\textbf{75.18} \rightarrow 69.73)$. This is likely due to decreased decoder capacity. 

\begin{figure*}[ht]
    \centering
    \includegraphics[width=\textwidth]{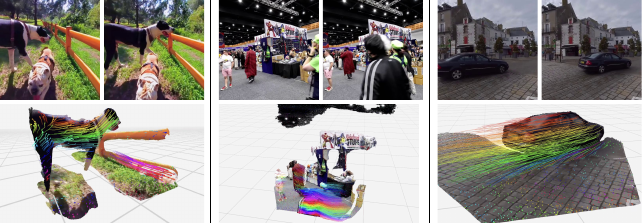}
    \caption{\textbf{More qualitative results on \dataset test set.} Extending~\Fig{result-wall-stereo4dtest}, we visualize image pairs and corresponding dynamic 3D point clouds predicted by DynaDUSt3R trained on \dataset. Our method recovers accurate 3D shape and complex scene motion.}
\label{fig:supp:qualitative-stereo4d}
\end{figure*}

\begin{figure*}[ht]
    \centering
    \includegraphics[width=\textwidth]{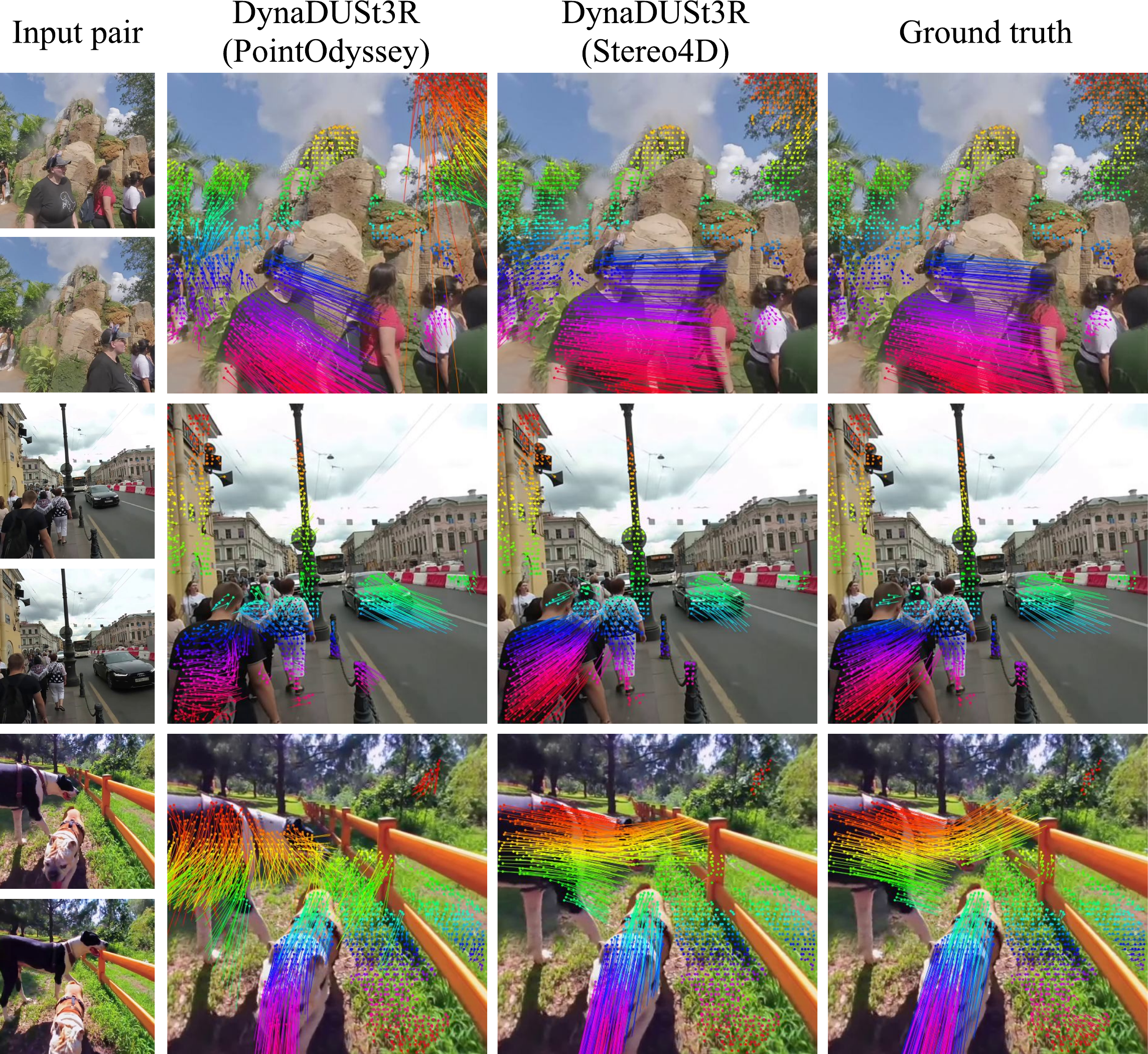}
    \caption{\textbf{More qualitative comparisons of 3D motion in the \dataset test set.} Extending~\Fig{compare-stereo4d}, we compare variants of DynaDUSt3R trained on different data sources. The Stereo4D-trained model also makes more precise motion predictions than the PointOdyssey-trained model.}
\label{fig:supp:compare-stereo4d}
\end{figure*}
\section{More qualitative comparisons}

\subsection{More results on held-out  \dataset examples}
\Fig{supp:qualitative-stereo4d} shows additional \method predictions on the \dataset held-out test set, extending \Fig{result-wall-stereo4dtest} from the main paper. \Fig{supp:compare-stereo4d} shows additional qualitative examples of motion comparisons on \dataset test set, extending~\Fig{compare-stereo4d} from the main paper. \Fig{supp:compare-stereo4d} compares variants of Dyna-DUSt3R trained on different data sources. The model trained on PointOdyssey incorrectly predicts large 3D motions, while the model trained on Stereo4D makes more accurate motion predictions, closer to ground truth.

\subsection{Additional track optimization examples}
In \Fig{supp:track_comparison}, we illustrate estimated tracks for a video sequence featuring a forward-moving camera and vehicles driving towards the camera. Our initial 3D tracks derived directly from RAFT depth, BootsTAP 2D tracks, and SfM camera pose, show significant jitter for both dynamic (vehicle) and static (ground) points. 
However, after applying our track optimization, the ground points produce stable, static tracks, and vehicle tracks become smooth and coherent.

\begin{figure*}[ht]
    \centering
    \includegraphics[width=\textwidth]{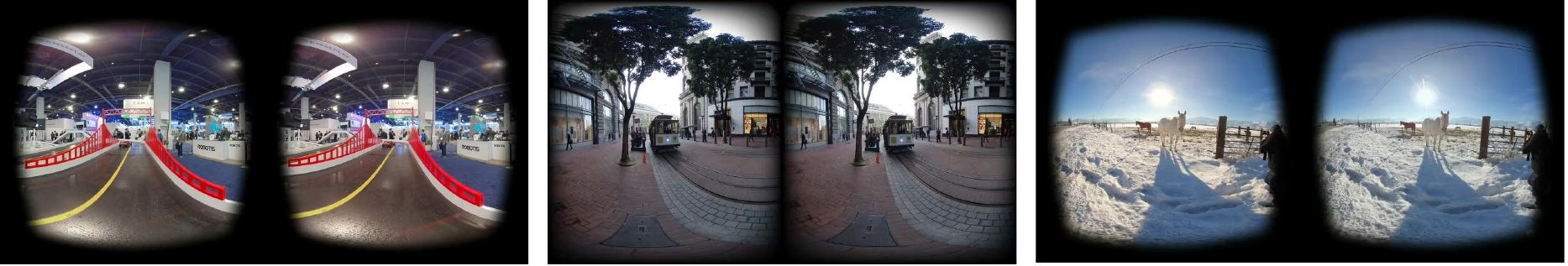}
    \caption{Example equirectangular stereo videos collected from the internet.}
    \label{fig:supp:equirect}
\end{figure*}

\begin{figure*}[ht]
    \centering
    \includegraphics[width=0.8\textwidth]{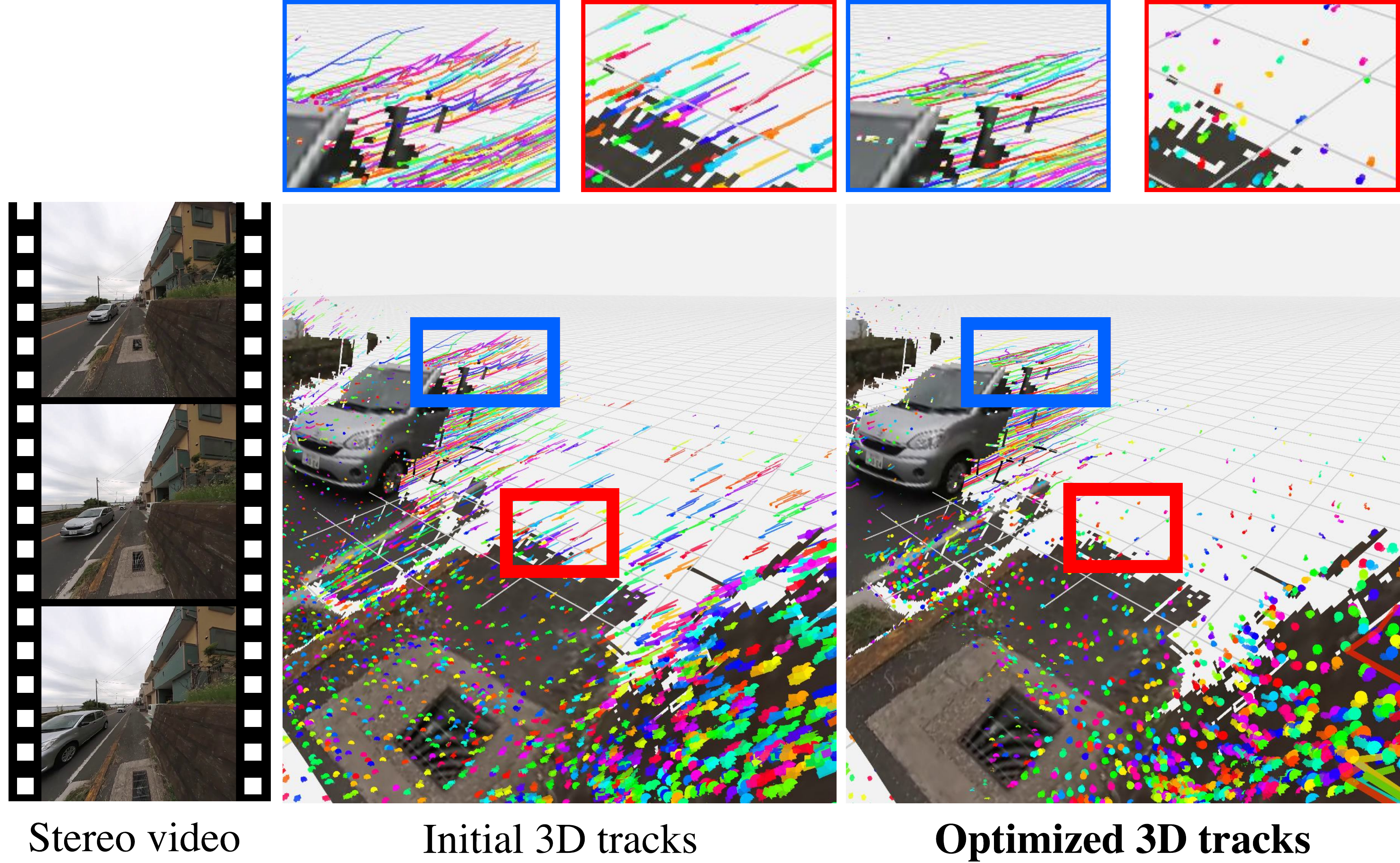}
    \caption{\textbf{Effect of Track Optimization.} We compare 3D tracks on a challenging walking tour video sequence. In this clip (left), the camera moves forward while vehicles drive toward the camera. We visualize the results across 16 frames, showing 3D trails left by both dynamic and static points.  \textbf{Middle}: Our initial 3D tracks, created directly from RAFT, BootsTAP and SfM camera pose, also exhibit significant jitter for both dynamic (vehicle) and static (ground) points.  \textbf{Right}: After applying our track optimization, the ground points yield stable, static tracks, and vehicle tracks become smooth and coherent.}
\label{fig:supp:track_comparison}
\end{figure*}

\section{Dataset curation details}
\subsection{Equirectangular videos}
The raw videos that we collect (see examples in \Fig{supp:equirect}) are natively stored in a cropped equirectangular format, which differs from a full 360$^\circ$ equirectangular projection as the horizontal field of view of the cropped format typically spans 180$^\circ$---half of a full sphere. These videos often contain metadata specifying the horizontal and vertical field of view. 
For instance, metadata for a typical video might specify 
$\mathsf{start_{yaw}}=-90.0^\circ$, $\mathsf{end_{yaw}}=90.0^\circ$,  $\mathsf{start_{tilt}}=-90.0^\circ$, $\mathsf{end_{tilt}}=90.0^\circ$; 
Since many VR180 videos are designed for an immersive VR experience, they are typically viewed with headsets. Hence, the baseline between the left and right cameras typically closely matches the average human interpupillary distance of 6.3 cm.

\subsection{Structure from motion}
For ease of processing with standard 3D computer vision pipelines, and to benefit from the wide FoV of the input videos, we convert the videos from their native format (equirectangular projections) to a fisheye format for camera pose estimation. 
We use a 140$^\circ$ field of view for these fisheye-projected videos, because many equirectangular videos have a black fade-out/feathering/vignetting effect applied at the boundary, as shown in~\Fig{supp:equirect}.
We found that using wider FoV frames significantly improves camera pose estimation in dynamic scenes. 
When using narrow FoV projections, dynamic objects are more likely to occupy a large fraction of the frame; when these dynamic foreground objects are rich in features, they can confuse camera tracking algorithms, leading to inaccurate camera poses that track the dynamic object rather than producing true camera motion with respect to the environment. 
In contrast, wide-angle fisheye videos capture more background regions, which tend to have stable features for tracking, yielding more reliable camera poses.

We first use ORB-SLAM2's stereo estimation mode~\cite{murartal2015orbslam}
to identify trackable sequences within the videos, utilizing the method devised by Zhou \etal to divide videos into discrete, trackable shots~\cite{zhou2018stereo}. 
For each given shot, consisting of frames $(I_i, \ldots, I_n)$, we estimate camera poses and rig calibration via an incremental global bundle adjustment algorithm similar to COLMAP~\cite{schonberger2016structure}. 
We initialize the stereo rig calibration to be that of a rectified stereo pair with baseline 6.3 cm, but optimize for the calibration as part of the bundle adjustment process, as in practice the stereo rig can vary significantly from its nominal configuration.
This process yields a camera position $\mathbf{c}_i$ and orientation $\mathbf{R}_i$ for each frame $i$ (defined as the pose of the left camera), and a position $\mathbf{c}_r$ and orientation $\mathbf{R}_r$ for the right camera relative to the left (assumed to be constant throughout the shot).

\subsection{Depth estimation}
Depth estimation is first performed on a per-frame basis, with disparity maps computed independently for each frame.  

We use the estimated camera rig calibration $\cB_r, \RB_r$ to rectify the original  high resolution equirectangular video frames, ensuring that (1) the left and right views have centered principal points, (2) are oriented perpendicular to the baseline, and (3) pointing in a parallel direction.  We then convert the equirectangular videos to  perspective projections for downstream predictions.

Disparity is estimated from optical flow~\cite{teed2020raft, sun2022disentangling} between the rectified left and right frames. 
The $x$-component of the optical flow is used as disparity, which is converted to metric depth using:
\begin{equation}
    \mathsf{Depth} = \frac{\mathsf{baseline}  \times f}{\mathsf{disparity}}.
\end{equation}
Here $\mathsf{baseline}=0.063$m, and $f$ is the frame's focal length.

\medskip
\noindent \textbf{Outlier Rejection.} Several criteria are applied to filter out unreliable pixels: 
\emph{Inconsistency between left and right eyes:} Disparity is rejected if the optical flow fails a cycle-consistency check with an error exceeding one pixel. 
\emph{Depth values exceeding 20 meters} are considered invalid. Estimating accurate depth beyond a certain range requires sub-pixel disparity estimation, and therefore the resulting depths are usually very noisy.
\emph{Negative flow values} that shouldn't occur, but can, often due to errors in textureless regions.
\emph{Large vertical flow:} pixels with a $y$-component of flow exceeding one pixel are removed (as in our rectified stereo pairs correspondences should have the same $y$-value, and violating that epipolar constraint indicates uncertain matches).
\emph{Occlusion boundaries:} Depth gradients exceeding a threshold ($\mathsf{threshold} = 0.3$) indicate occlusion boundaries and are rejected. For a pixel location $(x, y)$, depth gradients are computed as:
$$\mathsf{grad_x}=|{\mathsf{Depth}(x+1, y)-\mathsf{Depth}(x-1,y)} |,$$ $$\mathsf{grad_y}=|{\mathsf{Depth}(x, y+1)-\mathsf{Depth}(x,y-1)} |.$$
Pixels are rejected if $\mathsf{grad_x} > \mathsf{threshold} \times \mathsf{Depth}(x,y)$ or  $\mathsf{grad_y} > \mathsf{threshold} \times \mathsf{Depth}(x,y)$.

\subsection{2D tracks}
We extract long-range 2D point trajectories using Boots-TAP~\cite{doersch2024bootstap}. 
We run tracking on the left-eye video only. 
For every 10 frames, we uniformly initialize query points on image with stride 4. We then remove duplicated queries if earlier tracks fall within 1 pixel of a query point.

\subsection{Choice of FoV and resolution for perspective projection}
When converting the equirectangular videos to perspective projections, we use two FoVs: 60$^\circ$ and 120$^\circ$. Both perspective videos are set to a resolution of $512\times512$, the maximum supported by BootsTAP. The 60$^\circ$ projection offers a higher sampling rate in scene units, which improves the accuracy of depth estimation and 2D tracks when measured in meters. Additionally, it has smaller perspective distortion near the image boundaries. In contrast, the $120^\circ$ projection provides wider coverage, ensuring longer 2D tracks across the videos. This trade-off allows us to balance data quality with spatial coverage for downstream tasks, e.g. \method. We take the union of the 3D tracks derived from each of these videos for \method training supervision.

\section{\method training details}
\bfpar{Dataloader.} During training, we randomly sample two frames from the training videos that are at most 60 frames apart, at times $t_0$ and $t_1$, ($t_0 < t_1$). 
Additionally, we also sample one auxiliary frame in between, at time $t_{\mathsf{aux}}, t_0<t_\mathsf{aux}<t_1$, for additional track supervision between the two input frames. During training, we add data augmentation by applying random crops and color jitter to the input images and cropping the ground truth pointmap and motionmap accordingly. 

\bfpar{Training.} The network takes input the two RGB images as well as query times $t_q = \{0, 1, \frac{t_\mathsf{aux}-t_0}{t_1-t_0}\}$ and predicts the pointmaps for the two input views and motionmaps for each query $t_q$.
We supervise the network with losses defined in Eqn.~\ref{eqn:loss_point} and \ref{eqn:loss_motion}. We initialize our network with the \duster weights and initialize the motion head with the same weights as the point head. We finetune for 49k iterations with batch size 64, learning rate $2.5\times 10^{-5}$, and optimized by Adam with weight decay 0.05.

\end{document}